%% file: main.tex
\input{_constants}
\arxiv 

\pdfoutput=1
\documentclass[10pt,twocolumn,letterpaper]{article}
\input{cvpr_header}
\begin{document}
\title{\paperTitle}
\author{\authorBlock}
\maketitle

\input{00_abstract}
\input{01_intro}

\input{02_related}

\input{03_method}
\input{04_experiments}

\input{10_conclusion}

\noindent\textbf{Acknowledgements.}
{\small This paper has been supported by the French National Research
Agency (ANR) in the framework of its JCJC 
This work was granted access to the HPC resources of IDRIS under the allocation AD011013071 made by GENCI.

}

{\small
\bibliographystyle{ieee_fullname}
\bibliography{references}
}

\clearpage

\input{12_appendix}


\end{document}

%% file: _constants.tex
\def\confName{CVPR}
\def\confYear{2023}

\def\paperTitle{One-shot Unsupervised Domain Adaptation with Personalized Diffusion Models}

\def\authorBlock{
    Yasser Benigmim\textsuperscript{1,2}\quad  
    Subhankar Roy\textsuperscript{1} \quad
    Slim Essid\textsuperscript{1}\quad
    Vicky Kalogeiton\textsuperscript{2} \quad
    Stéphane Lathuilière\textsuperscript{1} \\
    \textsuperscript{1}\,LTCI, Télécom-Paris, Institut Polytechnique de Paris \\
    \textsuperscript{2}\,LIX, Ecole Polytechnique, CNRS,  Institut Polytechnique de Paris\\
    {\tt\small yasser.benigmim@telecom-paris.fr}
}

\newif\ifreview 
\newif\ifarxiv \newcommand{\arxiv}{\arxivtrue}
\newif\ifcamera 
\newif\ifrebuttal 

%% file: cvpr_header.tex
\ifreview \usepackage[review]{cvpr} \fi
\ifarxiv \usepackage[pagenumbers]{cvpr} \fi
\ifrebuttal \usepackage[rebuttal]{cvpr} \fi
\ifcamera \usepackage{cvpr} \fi

\usepackage{graphicx}
\usepackage{amsmath}
\usepackage{amssymb}
\usepackage{booktabs}

\input{_macros}  

\usepackage{xr-hyper}

\makeatletter
\newcommand*{\addFileDependency}[1]{
  \typeout{(#1)}
  \@addtofilelist{#1}
  \IfFileExists{#1}{}{\typeout{No file #1.}}
}

\makeatother

\usepackage[pagebackref,breaklinks,colorlinks]{hyperref}
\usepackage[capitalize]{cleveref}
\crefname{section}{Sec.}{Secs.}
\crefname{table}{Table}{Tables}
\crefname{figure}{Fig.}{Figs.}

%% file: _macros.tex
\usepackage{times}
\usepackage{microtype}
\usepackage{epsfig}
\usepackage[dvipsnames,xcdraw]{xcolor}
\usepackage{caption}
\usepackage{float}
\usepackage{placeins}
\usepackage{color, colortbl}
\usepackage{stfloats}
\usepackage{enumitem}
\usepackage{tabularx}
\usepackage{xstring}
\usepackage{multirow}
\usepackage{xspace}
\usepackage{xcolor}
\usepackage[hang,flushmargin]{footmisc}
\usepackage{booktabs}
\usepackage{graphicx}
\usepackage[export]{adjustbox}
\usepackage{url}
\usepackage{makecell}
\usepackage{math}
\usepackage{subcaption}
\usepackage{soul}

\usepackage[utf8]{inputenc}
\usepackage{pgfplots}
\DeclareUnicodeCharacter{2212}{−}
\usepgfplotslibrary{groupplots,dateplot}
\usetikzlibrary{patterns,shapes.arrows}

\pgfplotsset{compat=newest}
\usepgflibrary{plotmarks} 

\ifcamera \usepackage[accsupp]{axessibility} \fi

% ...

\ifarxiv  \fi

\newcommand{\R}[1]{{%
    \textbf{%
        \ifstrequal{#1}{1}{\textcolor{red}{R#1}}{%
        \ifstrequal{#1}{2}{\textcolor{blue}{R#1}}{%
        \ifstrequal{#1}{3}{\textcolor{magenta}{R#1}}{%
        \ifstrequal{#1}{4}{\textcolor{teal}{R#1}}{%
                           \textcolor{cyan}{R#1}%
        }}}}%
    }%
}}

\newcommand{\deeplab}{GreenYellow!30}
\newcommand{\segfor}{Gray!30}

\newcommand{\none}{Tan!30}
\newcommand{\all}{GreenYellow!30}
\newcommand{\one}{Gray!30}

\newcommand{\gtoc}{GTA $\to$ Cityscapes\xspace}
\newcommand{\stoc}{SYNTHIA $\to$ Cityscapes\xspace}
\newcommand{\task}{UDA\xspace}
\newcommand{\dm}{DM\xspace}
\newcommand{\daf}{DAFormer\xspace}
\newcommand{\osda}{OSUDA\xspace}

\newcommand{\method}{DATUM\xspace}
\newcommand{\fsda}{FSDA\xspace}
\newcommand{\db}{DreamBooth\xspace}
\newcommand{\ts}{\#TS\xspace}
\newcommand{\miou}{mIoU\xspace}
\newcommand{\methodname}{Data Augmentation with Diffusion Models}
\newcommand{\minus}[1]{\cellcolor{Cyan!#1}}
\newcommand{\plus}[1]{\cellcolor{BrickRed!#1}}

\definecolor{cadmiumgreen}{rgb}{0.0, 0.42, 0.24}


%% file: 00_abstract.tex
\begin{abstract}
Adapting a segmentation model from a labeled source domain to a target domain, where a single unlabeled datum is available, is one of the most challenging problems in domain adaptation and is otherwise known as one-shot unsupervised domain adaptation (\osda). Most of the prior works have addressed the problem by relying on style transfer techniques, where the source images are stylized to have the appearance of the target domain. Departing from the common notion of transferring only the target ``texture'' information, we leverage text-to-image diffusion models (e.g., Stable Diffusion) to generate a synthetic target dataset with photo-realistic images that not only faithfully depict the style of the target domain, but are also characterized by novel scenes in diverse contexts. The text interface in our method \textbf{D}ata \textbf{A}ugmen\textbf{T}ation with diff\textbf{U}sion \textbf{M}odels (\method) endows us with the possibility of guiding the generation of images towards desired semantic concepts while respecting the original spatial context of a single training image, which is not possible in existing \osda methods. Extensive experiments on standard benchmarks show that our \method surpasses the state-of-the-art \osda methods by up to +7.1\%. The implementation is available at : \url{https://github.com/yasserben/DATUM}
\end{abstract}

%% file: 01_intro.tex
\section{Introduction}
\label{sec:intro}

Semantic segmentation (SS) is one of the core tasks in computer vision~\cite{chen2017deeplab,wang2020deep,xie2021segformer}, where a neural network is tasked with predicting a semantic label for each pixel in a given image~\cite{csurka2022semantic}. Given its importance, SS has received significant attention from the deep learning community and has found numerous applications, such as autonomous driving~\cite{chen2017importance,cakir2022semantic}, robot navigation~\cite{hurtado2022semantic}, industrial defect monitoring~\cite{reiss2021panda}.

The task of semantic segmentation is known to require pixel-level annotations which can be costly and impractical in many real-world scenarios, making it challenging to train segmentation models effectively. Moreover, the issue of \emph{domain shift}~\cite{torralba2011unbiased} can cause segmentation models to underperform during inference on unseen domains, as the distribution of the training data may differ from that of the test data.
\input{figs/teaser}
To make learning effective without needing annotations on the target domain, several Unsupervised Domain Adaptation (\task) methods have been proposed for the task of semantic segmentation~\cite{csurka2021unsupervised,vu2019advent,hoffman2018cycada,tranheden2021dacs}. 
Fundamentally, the \task methods collectively use the labeled (or \textit{source}) and the unlabeled (or \textit{target}) dataset to learn a model that works well on the target domain. Despite being impressive in mitigating the domain gap, the \task methods rely on the assumption that a considerably large dataset of unlabelled images is at disposal. 
However, collecting a large target dataset before adaptation poses as a bottleneck in the rapid adoption of segmentation models in real-world applications. 
To circumvent this issue, several works have investigated the feasibility of using just a small subset of the unlabeled target samples (at times just \textit{one} sample) to adapt the model.
This adaptation scenario is known as One-Shot Unsupervised Domain Adaptation (\osda)~\cite{benaim2018one,luo2020adversarial,gong2022one,wu2022style}, where, in addition to the source dataset, only a single \textit{unlabelled} target sample is available.

While the \osda setting is realistic and cost-effective, relying solely on a single target image makes it challenging for traditional \task methods to estimate and align distributions.
To address the lack of target data, the \osda approaches generally overpopulate the target dataset with source images \textit{stylized} as target-\textit{like} ones~\cite{luo2020adversarial,gong2022one}. 
Albeit effective, these methods result in a target dataset that is limited to the scene layouts and structures inherent to the source dataset (Fig.~\ref{fig:teaser} left).  
In this work, we argue that simply mimicking the style of the target is insufficient to train a robust target model, especially when only limited information about the target domain is available.
Thus, we seek for diversifying the scene content and spatial layout, more than what the source images can offer.
Moreover, generating high-fidelity images is yet another challenging problem.
Thus, in this work, we focus on denoising diffusion models (\dm)~\cite{ho2020denoising,rombach2022high}, a family of generative models with excellent capability in generating high-quality images. 
We propose to leverage DMs to augment the target dataset with images that not only resemble the target domain at hand, but also contain diverse and plausible scene layouts due to rich prior knowledge encoded in DMs (see Fig.~\ref{fig:teaser} right).

In detail, we fine-tune a DM~\cite{ho2020denoising,rombach2022high} on the single target sample to generate an auxiliary large target dataset.
Following recent work~\cite{ruiz2022dreambooth,gal2023image}, we represent the target image with a special, rare and unique \textit{token} that encapsulates its visual appearance.
Then, we exploit the vast knowledge of DMs about the objects (or \textit{things} classes) present in the source domain for a driving scenario \cite{saharia2022photorealistic,liu2022compositional,fan2022frido}. Specifically, we prompt the model to generate a target dataset depicting such objects in a multitude of scenes, while maintaining the appearance tethered to the overall target domain \textit{style} via the unique token.
Once an augmented target dataset is made available, any \task  method can be used to adapt to the target domain.
We thus present our method \textbf{D}ata \textbf{A}ugmen\textbf{T}ation with diff\textbf{U}sion \textbf{M}odels (\textbf{\method}), for addressing \osda, as a connotation to the setting of having access to a single ``datum'' from the target domain. 
Our approach has the advantage of making any \task method compatible with the one/few-shot setting. 
In our experiments, we add \method to 
existing UDA methods and compare against the state-of-the-art \osda. Our results and analysis demonstrate the efficacy of \method and its ability to diversify the target dataset.
We believe that \method can contribute significantly to semantic segmentation as a \emph{plug-and-play} module.

Our \textbf{contributions} are three-fold: (\textbf{i}) We demonstrate, for the first time in the context of SS, the importance of generating semantically diverse and realistic target-like images in \osda.
(\textbf{ii}) We propose \method, a generic data augmentation pipeline powered by DMs, for addressing the challenging yet relevant task of \osda, and (\textbf{iii}) while being conceptually simple, we show with extensive experiments, on standard sim-to-real \task benchmarks, that \method can easily surpass the state-of-the-art \osda methods.

%% file: figs/teaser.tex
\begin{figure}[tp]
\centering
    \includegraphics[width=0.85\linewidth]{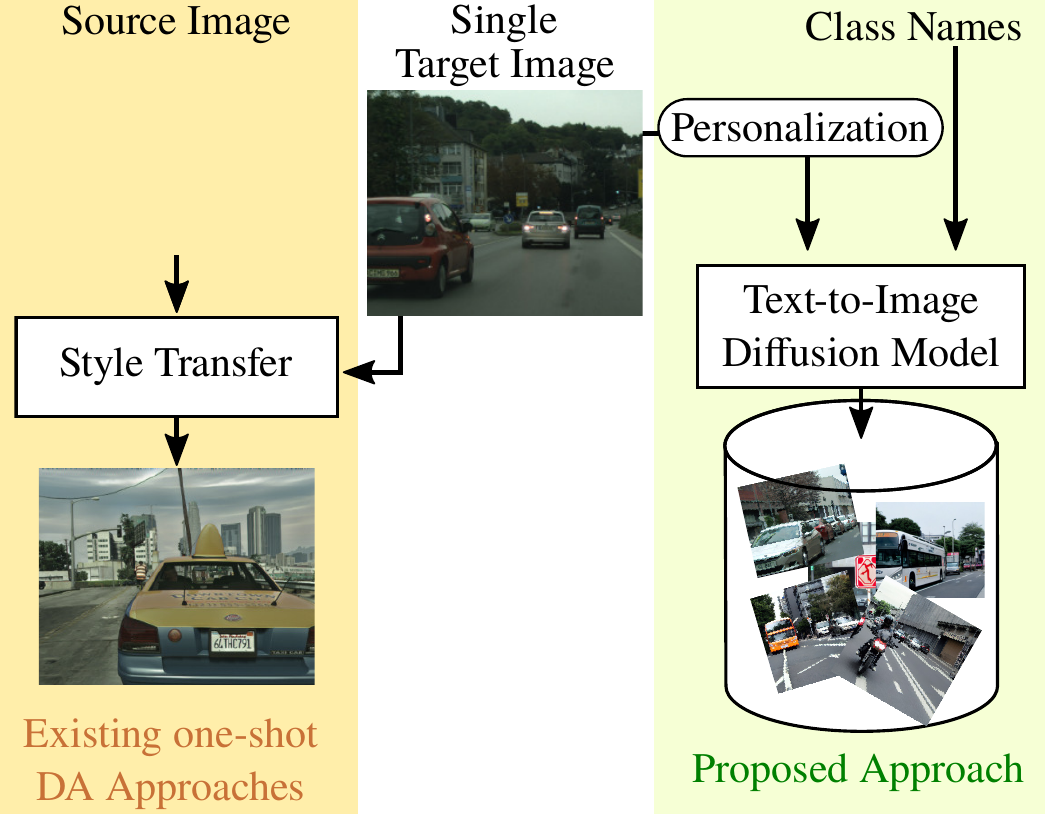}
    
    \caption{In existing \osda methods data augmentation is done via stylization \cite{gong2022one,luo2020adversarial}. In our proposed approach, we prompt the text-to-image diffusion models~\cite{rombach2022high} to generate new images that not only depict the style of the target domain, but also more faithfully capture the diversity of the scene content. 
    }
    \label{fig:teaser}
    \vspace{-3mm}
\end{figure}

%% file: 02_related.tex
\section{Related Works}
\label{sec:related}

\noindent\textbf{Unsupervised domain adaptation.}
To bridge the domain gap between the source and target datasets, unsupervised domain adaptation (\task) methods have been proposed, which can be roughly categorized into three broad sub-categories depending on the level where the distribution \textit{alignment} is carried out  in the network pipeline. First, the feature-level alignment methods aim at reducing the discrepancy between the source and target domains in the latent feature space of the network under some metric. As an example, these methods include minimizing the Maximum Mean Discrepancy (MMD)~\cite{bermudez2018domain} or increasing the domain confusion between the two domains with a discriminator network~\cite{huang2018domain,hoffman2016fcns,shen2019regularizing,vu2019advent,luo2019taking}. The latent space being high dimensional, the second category of \task methods~\cite{tsai2018learning,tsai2019domain,pan2020unsupervised,vu2019advent} exploits the lower dimensional and more structured output space of the network to mitigate domain shift, while borrowing \eg, adversarial alignment techniques. The third category includes methods~\cite{hoffman2016fcns,huang2018domain,hoffman2018cycada,shen2019regularizing,vu2019advent,luo2019taking,li2019bidirectional,toldo2020unsupervised} that align the source and the target domains in the input (or pixel) space by generating target-like source images via style transfer~\cite{gatys2016neural,huang2017arbitrary,zhu2017unpaired}. There is yet another successful line of UDA works that exploit self-training using a student-teacher framework~\cite{hoyer2022daformer,hoyer2022hrda,araslanov2021self}.

While the above \task methods are effective under the standard adaptation setting to varying degrees, where the entire target dataset is available for training, style transfer-based methods are particularly effective when the target data is inadequate to approximate a distribution. 
Different from the existing methods~\cite{benaim2018one,gong2022one}, which are just capable of transferring style (or ``appearance'') information to the source images, our proposed \method can additionally generate novel and structurally coherent content in the target domain.

\vspace{1mm}

\noindent\textbf{Few-shot adaptation.}
To improve the sample efficiency of the (UDA) methods, \textit{supervised} few-shot domain adaptation (\fsda) methods~\cite{motiian2017few,dong2018few,wang2019few} relax the need of having a large unlabeled target dataset, in favour of assuming access to a few but \textit{labeled} samples of the target domain. The \fsda methods~\cite{zhao2021domain,yue2021prototypical} exploit the labeled target samples to construct prototypes to align the domains. The setting of \osda is a more challenging version of \fsda, where a \textit{single} target sample is available \textit{without} any annotation. Due to the lack of means of constructing prototypes or aligning distributions with a single target sample,  \osda methods~\cite{luo2020adversarial,wu2022style,gong2022one} are based on transferring style from the target sample to the source dataset to artificially augment the target dataset. Once augmented,  \task methods such as self-training~\cite{gong2022one}, consistency training~\cite{luo2020adversarial}, prototypical matching~\cite{wu2022style}, are applied. Similar to~\cite{gong2022one}, we use the self-training framework DAFormer~\cite{hoyer2022daformer} to adapt  to the generated target images. However, unlike the prior \osda works~\cite{luo2020adversarial,gong2022one,wu2022style}, \method's data generation pipeline is stronger, conceptually simpler and does not rely on many heuristics.

\vspace{1mm}

\noindent\textbf{Diffusion models.}
Very recently, diffusion models (DM)~\cite{sohl2015deep,ho2020denoising} have brought a paradigm shift in the generative modeling landscape, showing excellent capabilities at generating photo-realistic text-conditioned images~\cite{rombach2022high,saharia2022photorealistic,ramesh2022hierarchical}. To allow personalized and more fine-grained generation, works such as DreamBooth~\cite{ruiz2022dreambooth}, Textual Inversion~\cite{gal2023image} and ControlNet~\cite{zhang2023adding} have extended DMs with different levels of fine-tuning, offering more flexibility. However, a handful of recent works~\cite{akrout2023diffusion,he2023synthetic,sariyildiz2023fake} has explored the possibility of using a latent diffusion model~\cite{rombach2022high}, a fast alternative to DM, for generating class-conditioned \textit{synthetic} datasets, as replacements of the \textit{real} counterparts, to solve image recognition tasks.
In contrast to these approaches, we specifically address the problem of domain adaptation  by augmenting the target domain. We adopt a fine-tuning strategy~\cite{ruiz2022dreambooth} that explicitly incorporates the appearance of the target domain. Our approach associates a word identifier with the content of the target image, resulting in high-fidelity synthetic generations.

%% file: 03_method.tex
\section{Method}
\label{sec:method}

\input{figs/pipeline}

In this work, we propose \textbf{D}ata \textbf{A}ugmen\textbf{T}ation with diff\textbf{U}sion \textbf{M}odels (\textbf{\method}), a generic method for creating \textit{synthetic} target dataset by using a single real sample (and hence, \textit{one-shot}) from the target domain. 
The synthetic dataset is then used for adapting a segmentation model. Sec.~\ref{sec:prelim} introduces the task and gives a background about DM, while Sec.~\ref{sec:synthetic-data}  describes \method.

\subsection{Preliminaries}
\label{sec:prelim}

\noindent\textbf{Problem formulation.}
In this work, we address the problem of One-Shot Unsupervised Domain Adaptation (\osda), where we assume access to $N^\mathrm{S}$ labeled images from a source domain $D^\mathrm{S} = \{(X_{i}^\mathrm{S},Y_{i}^\mathrm{S})\}_{i=1}^{N^\mathrm{S}}$, where $X_{i}^\mathrm{S} \in \mathbb{R}^{H \times W \times 3}$ represents an RGB source image and $Y_{i}^\mathrm{S} \in \mathbb{R}^{H\times W \times |\mathcal{C}|}$ the corresponding one-hot encoded ground-truth label, with each pixel belonging to a set of $\mathcal{C}$ classes. Unlike, traditional UDA methods~\cite{hoffman2018cycada,vu2019advent}, in \osda we have have access to a \textit{single unlabeled} target sample $X^\mathrm{T}$, where $X^\mathrm{T} \in \mathbb{R}^{H \times W \times 3}$.

In the context of semantic segmentation, the goal in \osda is to train a segmentation model $f \colon \mathcal{X} \to \mathcal{Y}$ that can effectively perform semantic segmentation on images from the target domain. Given the sheer difficulty in training $f(\cdot)$ with the single target image, our method seeks to generate a synthetic target dataset by leveraging a text-to-image DM.

\vspace{1mm}
\noindent\textbf{Background on Diffusion Models.} 
Diffusion Models (DM)~\cite{ho2020denoising} approach image generation as an image-denoising task. We obtain a sequence of $T$ noisy images $X_1...,X_{T}$ by gradually adding random Gaussian noises $\epsilon_1...,\epsilon_{T}$ to an original training image $X_0$.
A parameterized neural network $\epsilon_\theta(\cdot, t)$ is trained to predict the noise $\epsilon_t$ from $X_{t}$ for every denoising step $t\in \{1,...,T\}$. Denoising is typically carried out with a U-Net~\cite{ronneberger2015u}.
To enable conditioning, the network $\epsilon_\theta(X_t, y, t)$ is conditioned on an additional input $y$. In the case of text conditioning, the embeddings from a text-encoder $\tau_\theta$ for the text $y$ are used to augment the U-Net backbone with the cross-attention mechanism~\cite{vaswani2017attention}. For a given image-caption pair, the conditional DM is learned using the following objective:

\vspace{-5mm}
\begin{equation}
    \label{eqn:ldm}
    \mathcal{L}_{DM} = \mathbb{E}_{X,y, \epsilon \sim \mathcal{N}(0, 1),t} \Big[|| \epsilon - \epsilon_\theta(X_t, t, \tau_\theta(y)) ||^2_2\Big] 
\end{equation}
To improve efficiency, we employ a DM, which operates in the latent space of a pre-trained autoencoder~\cite{rombach2022high}.

\subsection{\methodname}
\label{sec:synthetic-data}

Our proposed \method works in three stages and is shown in Fig.~\ref{fig:three graphs}. In the first stage, called the \textbf{personalization stage}, we fine-tune a pre-trained text-to-image DM model by using multiple crops from the single target image (see Fig.~\ref{fig:stage1}). This steers the DM towards the distribution of the target domain of interest. Next, in the second  \textbf{data generation stage}, we prompt the just fine-tuned text-to-image DM to generate a synthetic dataset that not only appears to be sampled from the target domain, but also depicts desired semantic concepts (see Fig.~\ref{fig:stage2}). Finally, the \textbf{adaptive segmentation} stage culminates the three stage pipeline of \method, where we combine the labeled source data with the synthetic pseudo-target data and train with a general purpose UDA method (see Fig.~\ref{fig:stage3}). Next, we describe each stage in detail.

\input{figs/method_qualitative}
 
\vspace{1mm}
\noindent\textbf{Personalization stage.} The goal of the personalization stage is to endow the pre-trained DM with generation capabilities that are relevant to the downstream task. This stage is crucial because simply generating out-of-domain photo-realistic images is not useful for the downstream task. As an example, as shown in Fig.~\ref{tab:qualitative_method}(b), when an out-of-the-box DM is prompted with $p$ = ``\textit{a photo of} [CLS]'', where CLS represents a user-provided object class from the dataset, the DM generates high-fidelity images that truly depict the desired semantic concept. However, when compared to the real target domain (see Fig.~\ref{tab:qualitative_method}(a)) the DM generated images of Fig.~\ref{tab:qualitative_method}(b) have little to no resemblance in appearance. Given that the labeled source dataset already provides a rich prior to the segmentation model about the object classes of interest, having more unrelated and unlabeled images is unappealing.

Thus, we strive to imprint the appearance of the target domain into the synthetic dataset, while just using a single real target sample, in order to obtain more targetted synthetic data. Towards that end, we use \db~\cite{ruiz2022dreambooth}, a recently proposed technique for fine-tuning the DM, that allows for the creation of novel images while staying faithful to the user-provided subset of images. In detail, \db associates a unique identifier $V_*$ to the subset of images as provided by the user
by fine-tuning the DM weights. Similarly, we fine-tune the DM on the single target image while conditioning the model with the prompt $p$ = ``\textit{a photo of} $V_*$ \textit{urban scene}''. This results in the unique identifier $V_*$ capturing the target domain appearance. Once trained, we prompt the fine-tuned DM with $p$ = ``\textit{a photo of} $V_*$ \textit{urban scene}'' and report the results in the Fig.~\ref{tab:qualitative_method}(c). We observe the stark improvement in the overall visual similarity with the reference target domain images depicted in Fig.~\ref{tab:qualitative_method}(a). 
As a result of the personalization step with $V_*$, we can now condition the DM to generate more samples of the desired target domain.

However, a thorough inspection of the generated images in Fig.~\ref{tab:qualitative_method}(c) reveals that the images lack diversity. The DM overfits to the single target image and loses its ability to generate many other objects.
For instance, some classes (such as \textit{car}) are repeated whereas others (such as \textit{bus}, \textit{bike}, \textit{truck}) never appear.  
To prevent this overfitting issue, we train the DM for a limited number of iterations. 
Moreover, we disable the class-specific \textit{prior-preservation} loss used in Dreambooth \cite{ruiz2022dreambooth}, designed for not forgetting other concepts, since our goal is to capture the essence of the target domain, rather than generating a desired object in many unrealistic and unnatural scenarios.
For fine-tuning the DM, we optimize the training objective described in Eq.~(\ref{eqn:ldm}).
\vspace{1mm}

\noindent\textbf{Data generation stage.}
In the post personalization stage, our goal is to generate a dataset of synthetic images of the target domain. As we use just a single target image in the personalization stage, the generation capability of the fine-tuned DM model can still be limited to few scenes. Therefore, to elicit diverse generations from the fine-tuned DM, at inference we use more targetted prompts than the ones used during training. Specifically, we employ class-wise prompts in the form of: ``\textit{a photo of a} $V_*$ [CLS]''. The [CLS] corresponds to the name of the \textit{things} classes (\eg, \textit{bus}, \textit{person}, etc.) we want to generate, as defined in~\cite{caesar2018coco}. Our choice of using only the ``things'' classes is motivated by the fact that in a driving application, the ``things'' classes mostly co-occur with ``stuff'' classes (\eg, \textit{building}, \textit{sky}). Thus, explicitly prompting the model to generate stuff classes is redundant.
As shown in Fig.~\ref{tab:qualitative_method}(d), injecting the ``things'' class names into the inference prompt leads to an improved diversity in the generations, while staying close to the target domain in appearance. This helps in combating the long-tailed phenomenon of the semantic segmentation datasets, where some minority classes (\eg, \textit{bike}) appear less frequently than others, such as \textit{cars}, and \textit{road}.

\vspace{1mm}
\noindent\textbf{Adaptive segmentation stage.} While the pseudo-target images in the synthetic dataset contain the user-desired object, they still lack pixel-level information. To overcome this limitation, we resort to UDA techniques that enable a segmentation model to be adapted to an unlabeled target dataset. In this work, we leverage UDA methods such as \daf~\cite{hoyer2022daformer} and HRDA~\cite{hoyer2022hrda}, but our approach is not exclusive to these two methods. Notably, the optimization objective of these two UDA methods remain unaltered. In summary, our proposed \method can transform any UDA method into an effective \osda method.

%% file: figs/pipeline.tex
\begin{figure*}
     \centering
     \begin{subfigure}[b]{0.35\textwidth}
         \centering
         \includegraphics[height=4cm]{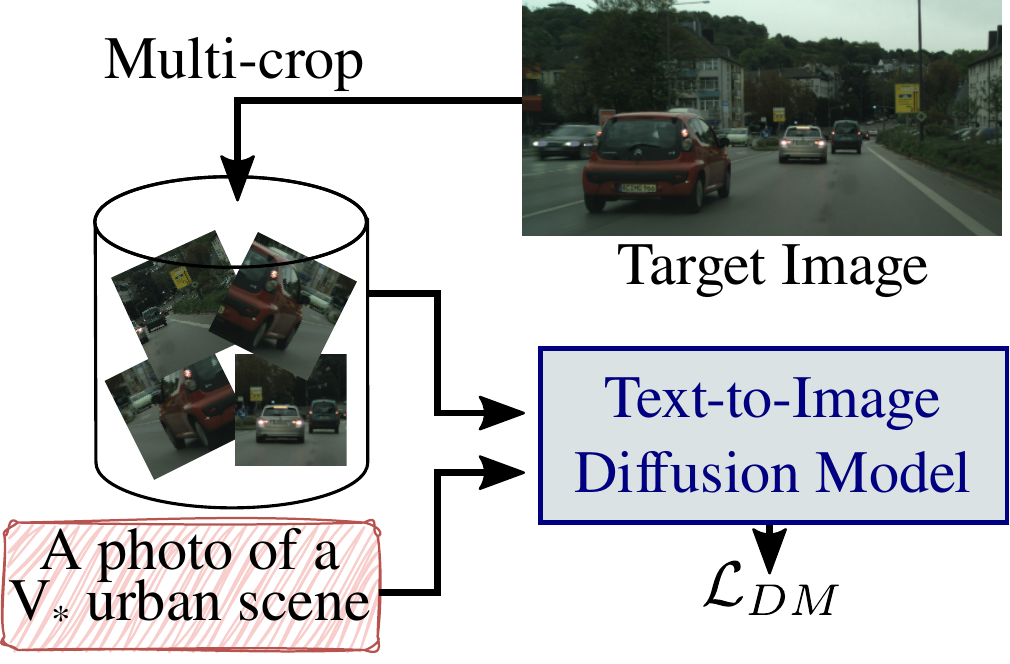}
         \caption{Personalization stage}
         \label{fig:stage1}
     \end{subfigure}
     \hfill
     \begin{subfigure}[b]{0.2\textwidth}
         \centering
         \includegraphics[height=4cm]{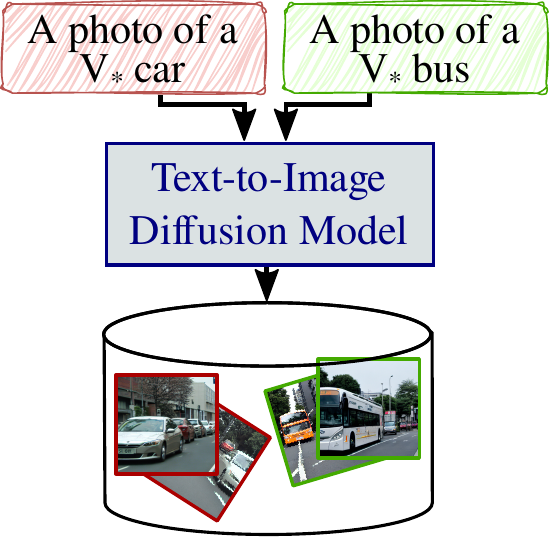}
         \caption{Data generation stage}
         \label{fig:stage2}
     \end{subfigure}
     \hfill
     \begin{subfigure}[b]{0.35\textwidth}
         \centering
         \includegraphics[height=4cm]{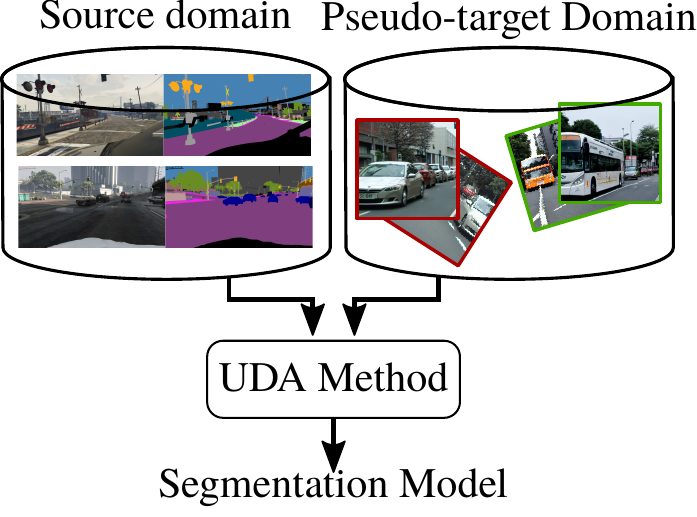}
         \caption{Adaptive segmentation stage}
         \label{fig:stage3}
     \end{subfigure}
        \caption{\textbf{The three stages of \method}. In the \textbf{personalization stage} (a), we learn to map a unique token $V_*$ with the appearance of the target domain using a single target image. In the \textbf{data generation stage} (b), we employ the personalized model to generate a large dataset corresponding to the target distribution. Class names are used to improve diversity. Finally, the \textbf{adaptive segmentation stage} (c) consists in training an existing UDA framework on the labeled source and the generated unlabeled pseudo-target datasets 
        }
        \label{fig:three graphs}
        \vspace{-4mm}
\end{figure*}

%% file: figs/method_qualitative.tex
\begin{figure*}[tp]
    \centering
    \setlength{\tabcolsep}{0.5pt}
    \resizebox{\textwidth}{!}{%
    \begin{tabular}{ccccc}
        \includegraphics[width=0.18\linewidth,height=44.7pt]{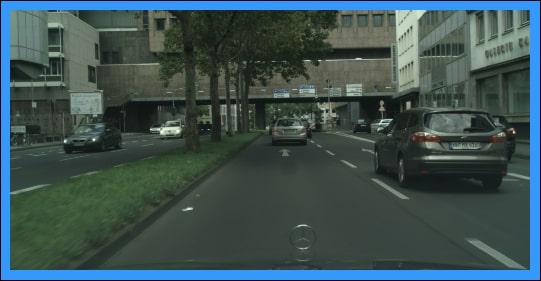} & \includegraphics[width=0.18\linewidth]{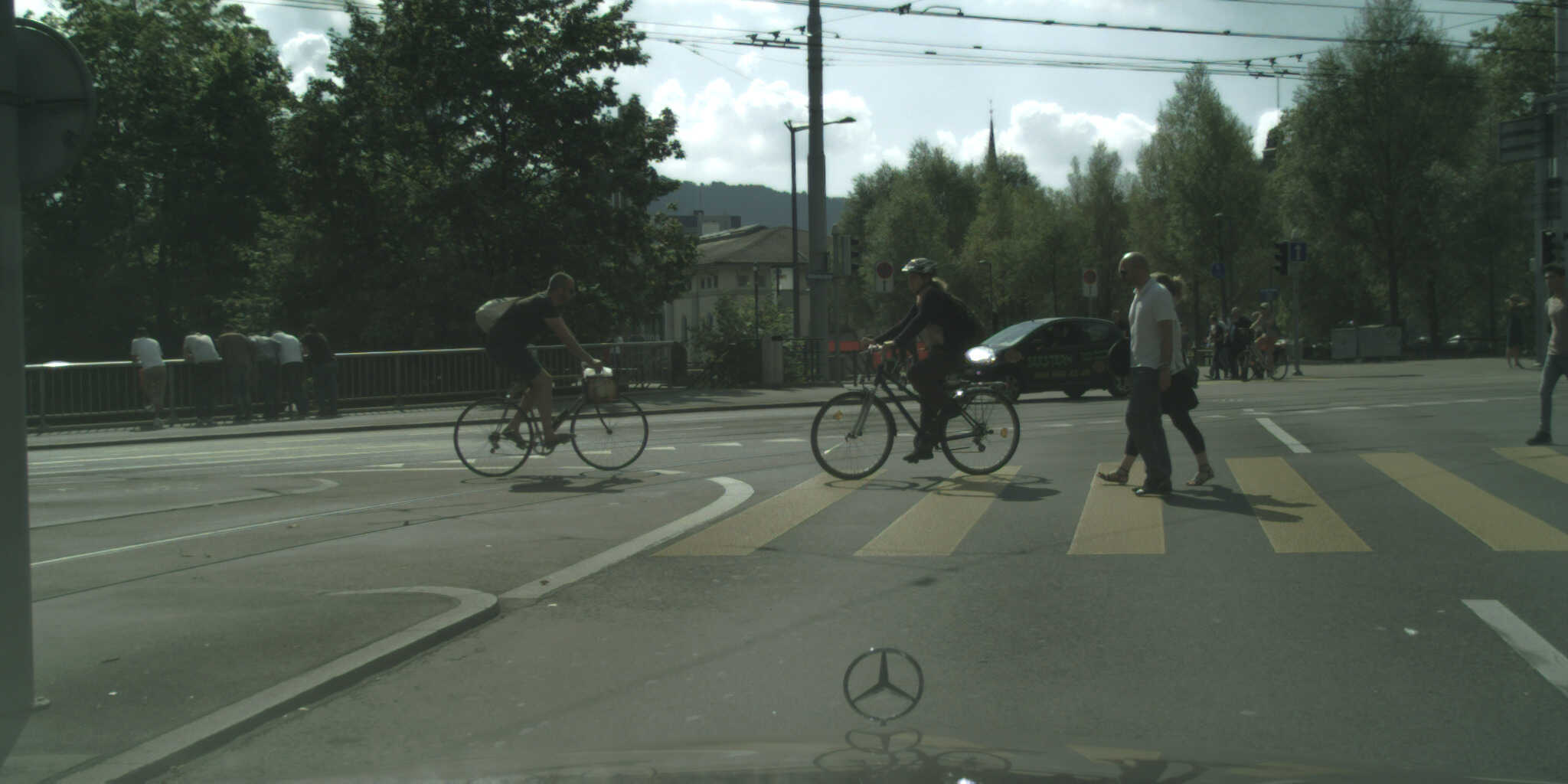} & \includegraphics[width=0.18\linewidth]{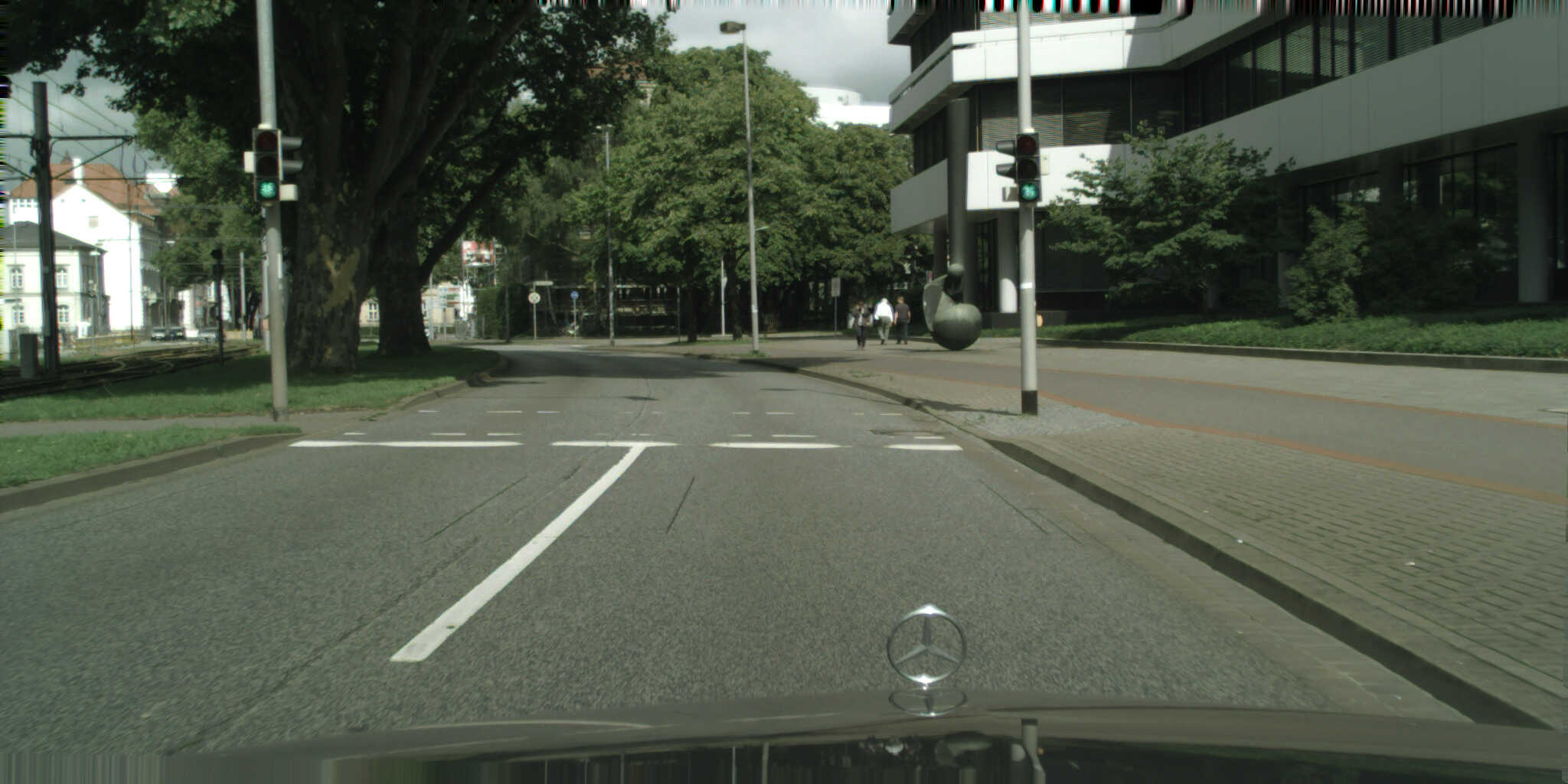} & \includegraphics[width=0.18\linewidth]{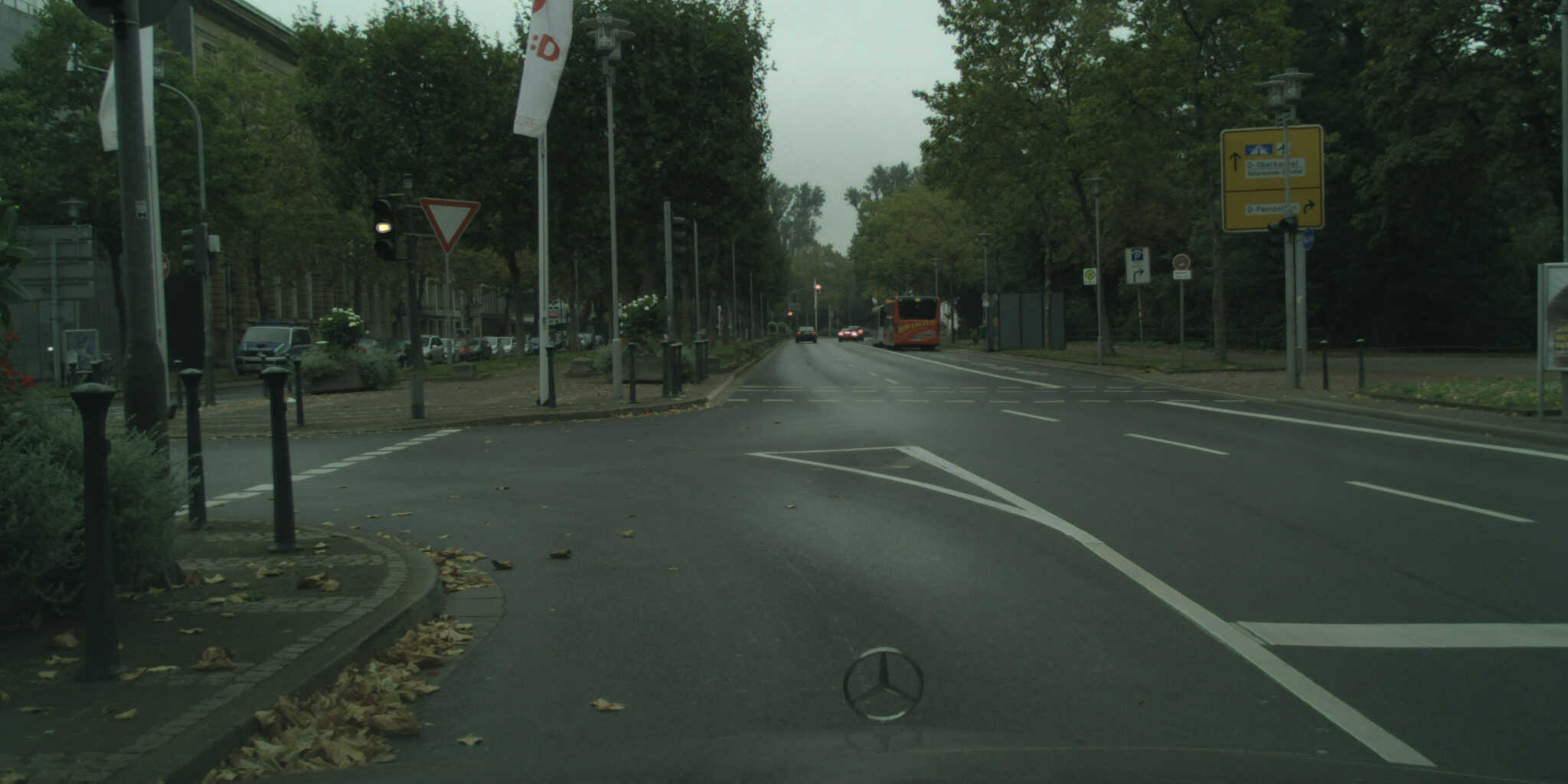} & \includegraphics[width=0.18\linewidth]{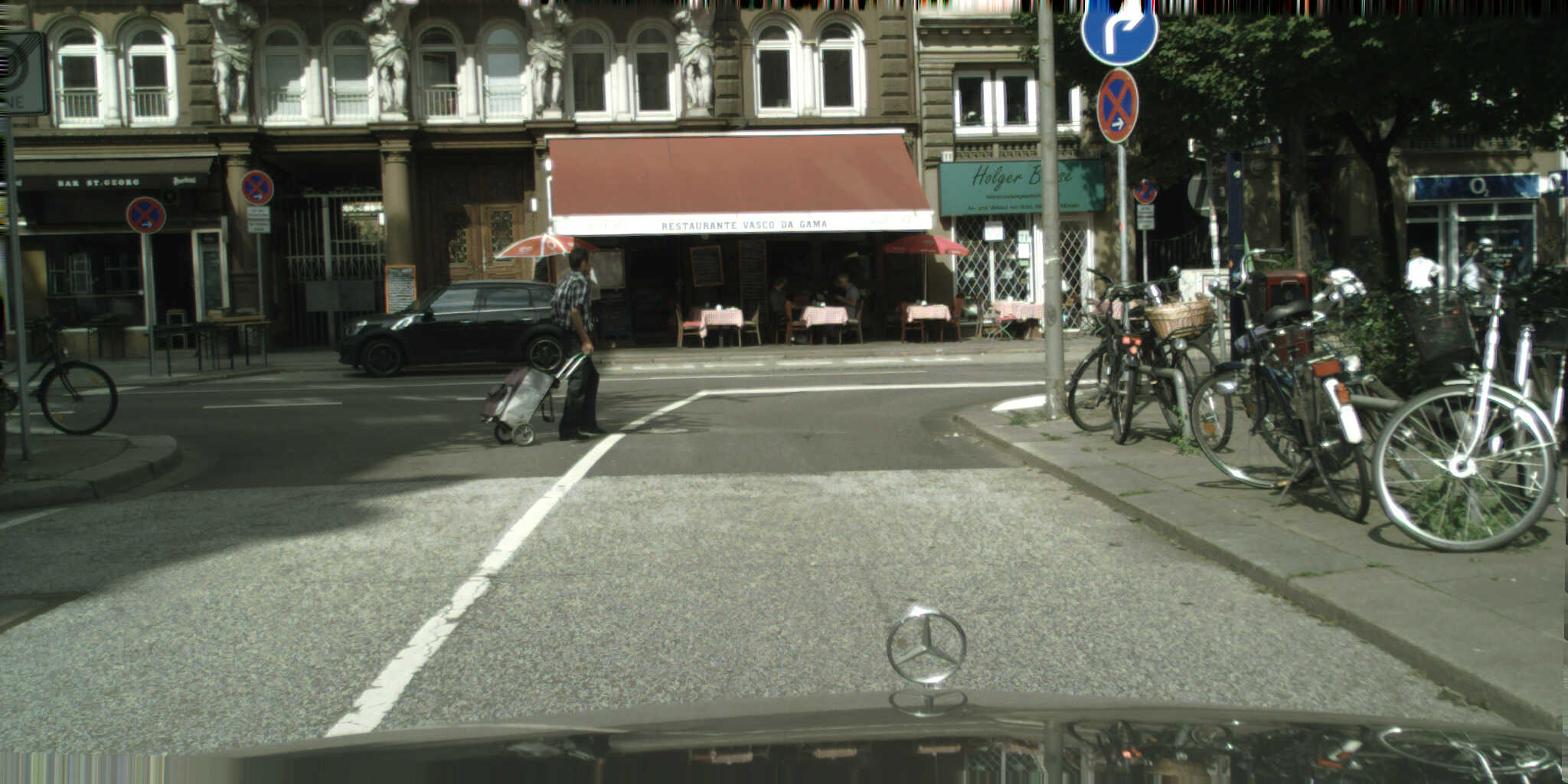} \\

        \multicolumn{5}{c}{\scriptsize (a) Real images from the target domain (Cityscapes) for reference} \\


        \begingroup  \sbox0{\includegraphics{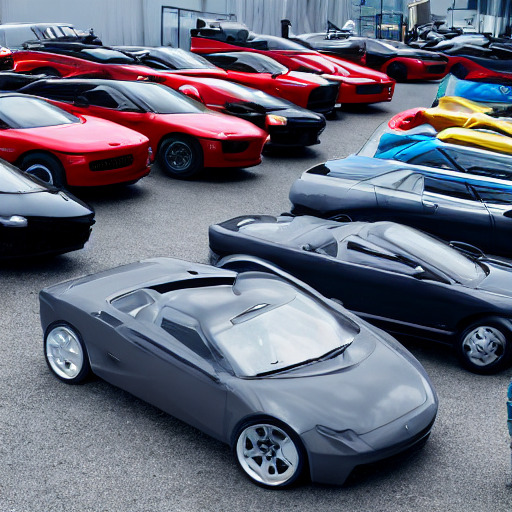}}%
        \includegraphics[clip,trim=0 {.4\wd0} 0 {.07\wd0},width=0.18\linewidth]{figs/assets/qualitative/stable_diffusion/img1.jpg}
        \endgroup
        
        & 

        \begingroup  \sbox0{\includegraphics{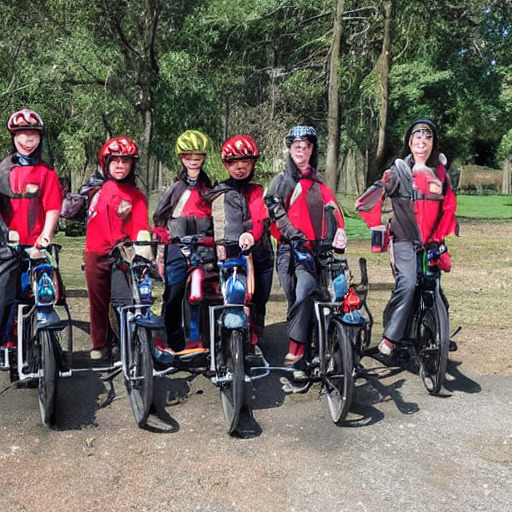}}%
        \includegraphics[clip,trim=0 {.25\wd0} 0 {.22\wd0},width=0.18\linewidth]{figs/assets/qualitative/stable_diffusion/img2.jpg}
        \endgroup
        
        & 

        \begingroup  \sbox0{\includegraphics{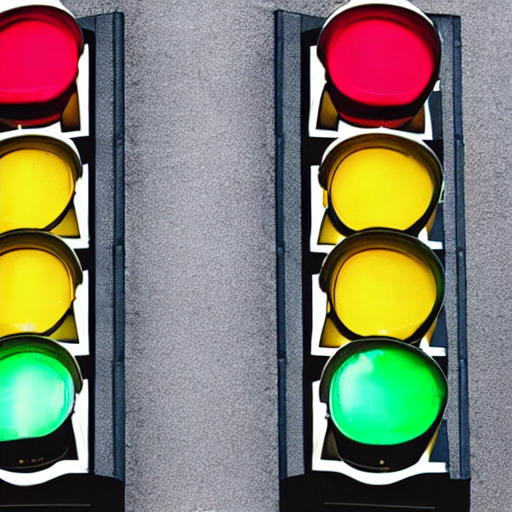}}%
        \includegraphics[clip,trim=0 {.47\wd0} 0 0,width=0.18\linewidth]{figs/assets/qualitative/stable_diffusion/img3.jpg}
        \endgroup
        
        & 

        \begingroup  \sbox0{\includegraphics{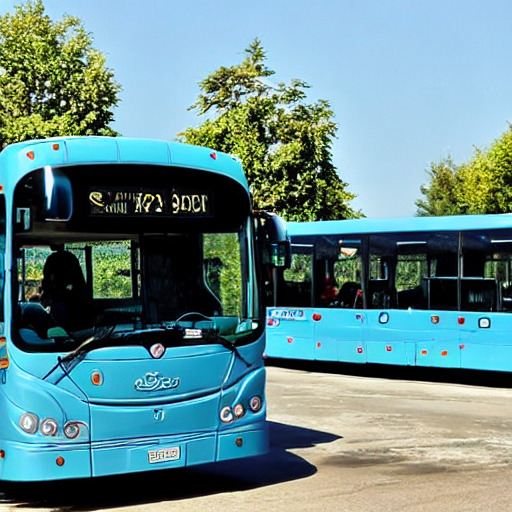}}%
        \includegraphics[clip,trim=0 {.20\wd0} 0 {.27\wd0},width=0.18\linewidth]{figs/assets/qualitative/stable_diffusion/img4.jpg}
        \endgroup
        
        & 

        \begingroup  \sbox0{\includegraphics{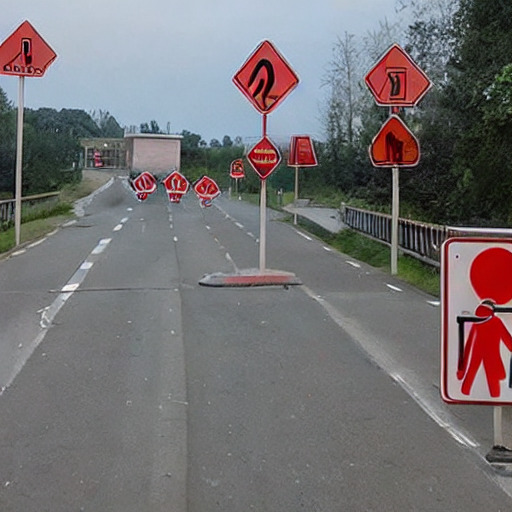}}%
        \includegraphics[clip,trim=0 {.35\wd0} 0 {.12\wd0},width=0.18\linewidth]{figs/assets/qualitative/stable_diffusion/img5.jpg}
        \endgroup \\

        \multicolumn{5}{c}{\scriptsize (b) Synthetic images from \textit{out-of-the-box} SD with the prompt $p$ = ``\textit{a photo of} [CLS]''
        } \\


        \begingroup  \sbox0{\includegraphics{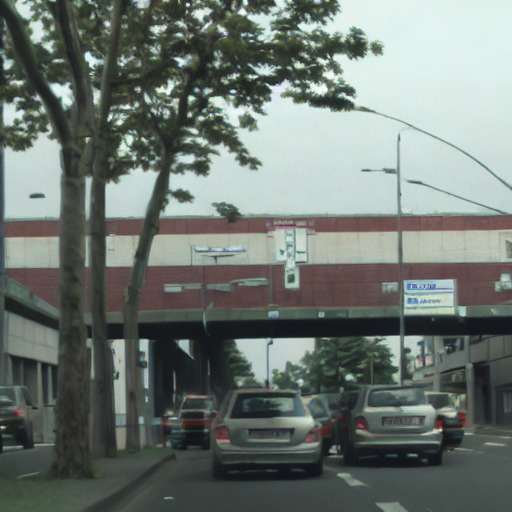}}%
        \includegraphics[clip,trim=0 {.07\wd0} 0 {.4\wd0},width=0.18\linewidth]{figs/assets/qualitative/dreambooth_v1_v1/img1.jpg}
        \endgroup
        
        & 

        \begingroup  \sbox0{\includegraphics{figs/assets/qualitative/stable_diffusion/img2.jpg}}%
        \includegraphics[clip,trim=0 {.17\wd0} 0 {.3\wd0},width=0.18\linewidth]{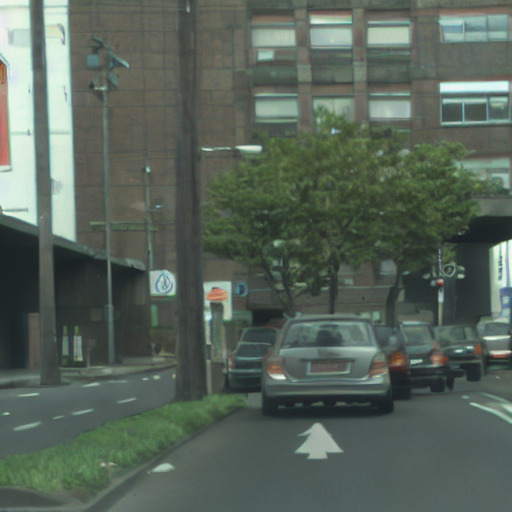}
        \endgroup
        
        & 

        \begingroup  \sbox0{\includegraphics{figs/assets/qualitative/stable_diffusion/img3.jpg}}%
        \includegraphics[clip,trim=0 {.1\wd0} 0 {.37\wd0},width=0.18\linewidth]{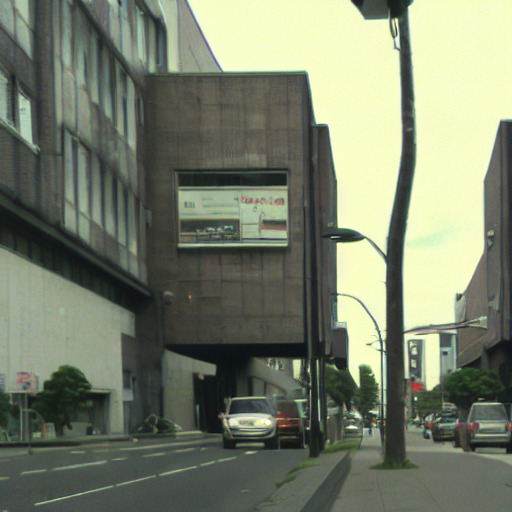}
        \endgroup
        
        & 

        \begingroup  \sbox0{\includegraphics{figs/assets/qualitative/stable_diffusion/img4.jpg}}%
        \includegraphics[clip,trim=0 {.22\wd0} 0 {.25\wd0},width=0.18\linewidth]{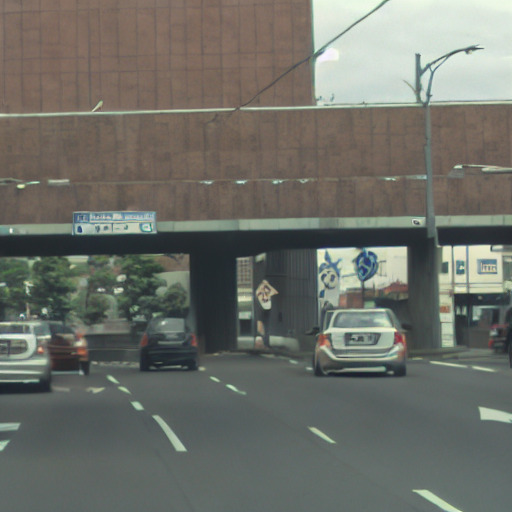}
        \endgroup
        
        & 

        \begingroup  \sbox0{\includegraphics{figs/assets/qualitative/stable_diffusion/img5.jpg}}%
        \includegraphics[clip,trim=0 {.2\wd0} 0 {.27\wd0},width=0.18\linewidth]{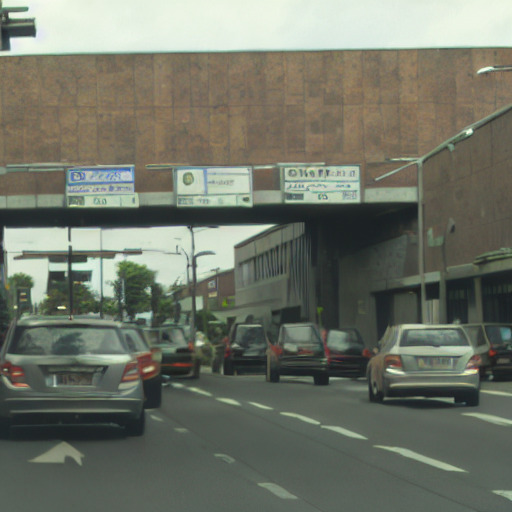}
        \endgroup \\

        \multicolumn{5}{c}{\scriptsize (c) Synthetic target images with the prompt $p$ = ``\textit{a photo of  $V_*$ urban scene}''}. \\


        \begingroup  \sbox0{\includegraphics{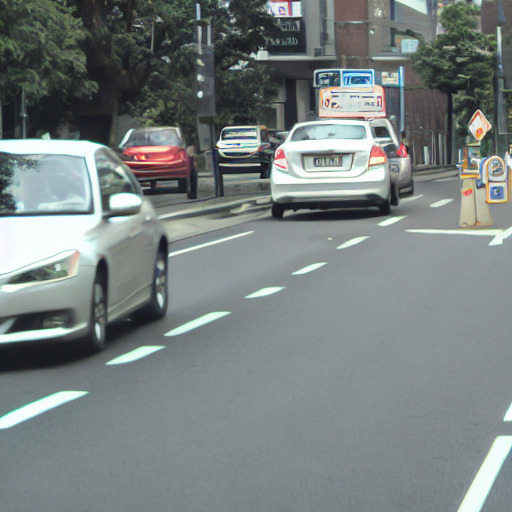}}%
        \includegraphics[clip,trim=0 {.35\wd0} 0 {.12\wd0},width=0.18\linewidth]{figs/assets/qualitative/dreambooth_v1_v2/img1.jpg}
        \endgroup
        
        & 

        \begingroup  \sbox0{\includegraphics{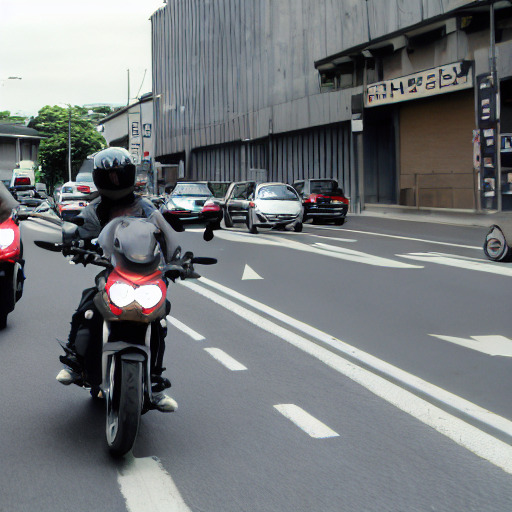}}%
        \includegraphics[clip,trim=0 {.24\wd0} 0 {.23\wd0},width=0.18\linewidth]{figs/assets/qualitative/dreambooth_v1_v2/img2.jpg}
        \endgroup
        
        & 

        \begingroup  \sbox0{\includegraphics{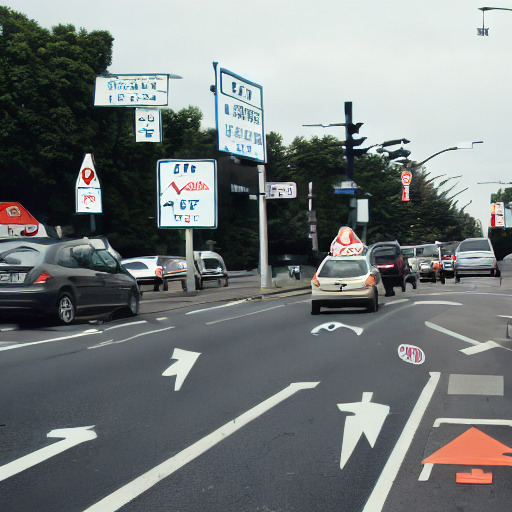}}%
        \includegraphics[clip,trim=0 {.34\wd0} 0 {.13\wd0},width=0.18\linewidth]{figs/assets/qualitative/dreambooth_v1_v2/img3.jpg}
        \endgroup
        
        & 

        \begingroup  \sbox0{\includegraphics{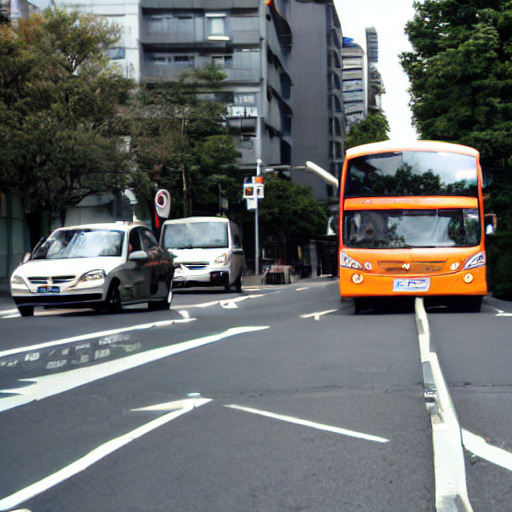}}%
        \includegraphics[clip,trim=0 {.34\wd0} 0 {.13\wd0},width=0.18\linewidth]{figs/assets/qualitative/dreambooth_v1_v2/img4.jpg}
        \endgroup
        
        & 

        \begingroup  \sbox0{\includegraphics{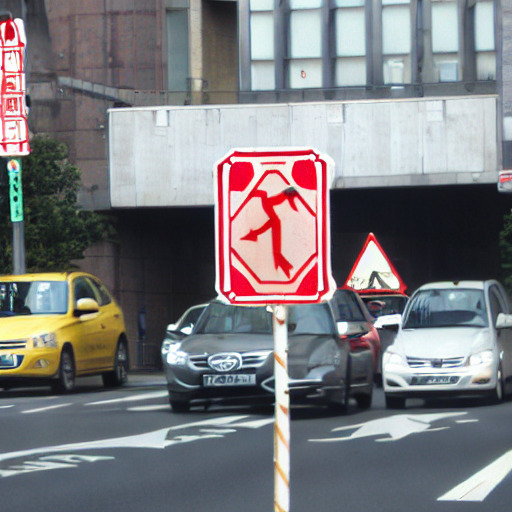}}%
        \includegraphics[clip,trim=0 {.2\wd0} 0 {.27\wd0},width=0.18\linewidth]{figs/assets/qualitative/dreambooth_v1_v2/img5.jpg}
        \endgroup \\

        \multicolumn{5}{c}{\scriptsize (d) Synthetic target images with the prompt $p$ = ``\textit{a photo of } $V_*$ [CLS]''} \\
        
    \end{tabular}%
    }
    \vspace{-3mm}
    \caption{\textbf{Qualitative study illustrating the underlying motivations of our three-stage approach}. (a) Real images from the Cityscapes target domain. (b) Out-of-the-box Stable Diffusion (SD) can generate photo-realistic images given the [CLS] name in the prompt, but barely have any resemblance to Cityscapes. (c) Fine-tuning SD  on a single target image (personalization stage) leads to generations that truly mimic the Cityscapes domain, but at the cost of losing diversity. (d) Our proposed prompting strategy  (data generation stage) leads to synthetic generations that are both photo-realistic and also ressembles Cityscapes-like images.The blue-framed image in (a) is the training image used to generate the images in rows (c) and (d).
    }
    \label{tab:qualitative_method}
    \vspace{-3mm}
\end{figure*}

%% file: 04_experiments.tex
\section{Experiments}
\label{sec:exps}

\subsection{Experimental set up}
\label{sec:setup}

\paragraph{Dataset and settings.}
We follow the experimental settings established in the \osda literature~\cite{luo2020adversarial,wu2022style,gong2022one} and conduct experiments on two standard \textit{sim-to-real} benchmarks: \gtoc and \stoc, where GTA~\cite{richter2016playing} and SYNTHIA~\cite{ros2016synthia} are the source domains in the respective settings, and Cityscapes~\cite{cordts2016cityscapes} is the target domain. In details, the GTA dataset comprises 24,966 synthetic images with a resolution of 1914 $\times$ 1052. and SYNTHIA contains 9400 synthetic images of resolution 1280 $\times$ 760. Cityscapes contains 2975 training images and 500 validation images of size 2048 $\times$ 1024, is captured under real-world driving conditions. Note that, since we operate in the one-shot adaptation scenario, as in~\cite{luo2020adversarial,wu2022style,gong2022one}, we assume to have access to a \textit{datum} from the target domain, which is chosen at random during training.

\vspace{1mm}

\noindent\textbf{Implementation details.}
We employ the Stable Diffusion (SD) implementation of Latent Diffusion Models (LDM)~\cite{rombach2022high}. We use the publicly available \textit{Diffusers} library~\cite{von-platen-etal-2022-diffusers} for all the experiments related to generating synthetic data. In particular, for generating synthetic images in the target domain, we start from the Stable Diffusion v1.4 checkpoint~\cite{sdv1_4} and fine-tune it using the DreamBooth~\cite{ruiz2022dreambooth} method. We refer the reader to LDM~\cite{rombach2022high} for details about the encoder, U-Net, and decoder architectures.

For fine-tuning SD, we randomly crop patches of 512 $\times$ 512 from the original 2048 $\times$ 1024 resolution, and use a generic prompt $p$ = ``\textit{a photo of a} $V_*$ \textit{urban scene}'', given that the target domain Cityscapes was captured in an urban set-up~\cite{cordts2016cityscapes}. We train SD for 200 iterations, and we find that for the one-shot setting longer training leads to overfitting on the target scene. Once trained, we generate a synthetic target dataset  of cardinality 2975, which is equivalent in size to the Cityscapes training set, by utilizing inference prompts of the form $p$ = ``\textit{a photo of a} $V_*$ [CLS]''.  \db generates images at the same resolution as the input, which is 512 $\times$ 512. This generated dataset then serves as the target domain for adaptation, as in \task.

For training the final segmentation model on the source and generated datasets, we use the network architecture from state-of-the-art \task methods~\cite{hoyer2022daformer} that use MiT-B5~\cite{xie2021segformer} as the encoder and a context-aware fusion~\cite{hoyer2022daformer} as the decoder. This is analogous to the most popular ResNet-101~\cite{he2016deep} as a backbone, and DeepLabV2~\cite{chen2017deeplab} as the decoder. 
We also experiment with another \task method:  HRDA~\cite{hoyer2022hrda}. For both these experiments with \daf and HRDA, we keep the training protocol and hyperparameters unchanged. Both the ResNet-101 and MiT-B5 are pre-trained on ImageNet-1k.

\vspace{1mm}
\noindent\textbf{Evaluation metrics.}
Following the standard protocol~\cite{luo2020adversarial}, we report the mean Intersection over Union (mIoU) on the validation set of Cityscapes. For the \gtoc benchmark, we compute mIoU over 19-classes, whereas for \stoc,  we report both mIoU\textsuperscript{13} and mIoU\textsuperscript{16} for 13 and 16 classes, respectively~\cite{gong2022one}.

\input{tables/g2c_sota}
\input{tables/s2c_sota}

\subsection{Comparison with the state-of-the-art}
\label{sec:sota}

\paragraph{Baselines.}
We compare our proposed method with the state-of-the-art \osda methods and \task methods adapted to the \osda setting: CycleGAN~\cite{zhu2017unpaired}, ASM~\cite{luo2020adversarial}, OST~\cite{benaim2018one}, CACDA~\cite{gong2022one}, SMPPM~\cite{wu2022style}, DACS~\cite{tranheden2021dacs} which are also methods based on data augmentation, as well as ProDA~\cite{zhang2021prototypical}, CBST~\cite{vu2019advent},    AdaptSeg~\cite{tsai2018learning}, DAFormer~\cite{hoyer2022daformer} and HRDA~\cite{hoyer2022hrda}. Given that \method focuses primarily on data generation, we pair it with the \task methods \daf and HRDA under the \osda setting.
We denote these models as \daf + \method and HRDA + \method, which use \textbf{purely synthetic} target dataset generated by \method, alongside source.
For a fair comparison with baselines, we use the DAFormer network architecture (with MiT-B5 backbone), which has demonstrated superior effectiveness compared to weaker counterparts, such as ResNet-101~\cite{hoyer2022daformer}. However, as performance metrics for some older \osda methods~\cite{benaim2018one,luo2020adversarial} are not available with a \daf-like architecture, we also experiment using DeepLabV2 with ResNet-101 backbone.

\vspace{1mm}
\noindent\textbf{Main results.}
In Tab.~\ref{tab:g2c-sota} and Tab.~\ref{tab:s2c-sota} we report the results on the \gtoc and the \stoc benchmarks, respectively, under the traditional UDA as well as the \osda setting.
The traditional UDA setting~\cite{hoyer2022daformer} is  denoted as \colorbox{\all}{All}, as it uses all target samples, while the \osda setting is denoted as \colorbox{\one}{One} since we have access to only a single datum.
Following the standard practices from the one/few shot learning literature, we report our results averaged over 3 independent runs using randomly sampled unlabeled real target datum. Also note that in our experiments we report the model performance after the last training iteration, instead of picking the maximum \miou.

From the Tab.~\ref{tab:g2c-sota} we notice that using our generated target dataset for training the state-of-the-art \task methods in the \osda setting, greatly improves their performances, independent of the backbone. For instance, \daf + DATUM is +6.4\% better (43.2 $\to$ 49.6\%) than \daf, with the ResNet-101 as backbone. Similar trends can be noticed 
when using the MiT-B5 backbone, where we improve HRDA by +8.3\% (\ie, from 52.9\% $\to$ 61.2\%). Overall, for the \gtoc with the MiT-B5 as backbone, we beat the best competitor CACDA~\cite{gong2022one} by significant margins (55.4\% versus our 61.2\%). Interestingly, we observe that while using the ResNet-101 backbone, our data generation can even outperform UDA methods that use all the original target dataset, \eg, CycleGAN and ASM.

From Tab.~\ref{tab:s2c-sota}, which reports the performance on the \stoc benchmark, we observe similar results. Pairing our generated dataset with \task methods consistently improves performance under the \osda setting. Compared to the best competitor method CACDA, \method helps achieve the new state-of-the-art results, by comprehensively outperforming CACDA by +2.0\% and +3.1\% in  mIoU\textsuperscript{13}.
We believe that these findings are highly significant in bridging the gap between \osda and standard UDA.

\input{tables/ablation/style_transfer}

\vspace{1mm}
\noindent\textbf{Comparison with style-transfer methods.}
Given that \method is akin to data augmentation in image stylization, we compare it against two style transfer techniques used in existing \osda methods:  RAIN~\cite{luo2020adversarial} and PT+CDIR~\cite{gong2022one}. Tab.~\ref{tab:style_transfer}  reports the results. We observe that generating novel scenes with \method is more impactful than simply augmenting the source images with the target style as in the other methods.

\subsection{Ablation analysis}
\label{sec:ablation}
To examine the effectiveness of \method, in this section we conduct thorough ablation analyses of each component associated with it. All ablations are carried out on \gtoc benchmark with \daf~\cite{hoyer2022daformer} using only one random datum from the target domain.

\input{figs/ablation_1_2}

\vspace{1mm}
\noindent\textbf{Impact of number of shots.}
To investigate the impact of the number of real target samples (or \ts) on the \osda performance, we conduct an ablation study where we vary the \ts and personalize SD with \method for a varied number of training iterations. In Fig.~\ref{fig:num-shots-miou} we plot the performance of \daf + \method for different \ts and compare it with SD. We observe that for lower \ts (\ie, one shot) \method achieves the best performance, and the \miou gradually degrades with prolonged training. This is because the SD overfits on the single target image and loses its ability to generate diverse scenes. The issue is less severe when the \ts increases to 10 (\ie, ten shot), and the \miou is fairly stable. Nevertheless, \method generates more informative target images than SD, highlighting the need for incorporating the target style into the synthetic dataset generation process.

\input{tables/ablation/prompts}

\vspace{1mm}
\noindent\textbf{Impact of prompts.}
Since \method depends on the choice of prompts used during training and inference, here we ablate the impact of training and inference prompts by quantitatively measuring the \miou for different combinations and report the results in Tab.~\ref{tab:prompts}. We observe that the combination of training prompt $p$ = ``\textit{a photo of a} $V_*$ \textit{urban scene}'' and class-aware inference prompt $p$ = ``\textit{a photo of a} $V_*$ [CLS]'' leads to the best results (second row). 
When compared to the class agnostic inference prompt $p$ = ``\textit{a photo of a} $V_*$ \textit{urban scene}'' (first row), the performance increases by +4.3\% . This demonstrates that grounding \method with things/objects of interest leads to more meaningful scene composition, and provides more information to the segmentation model. Using the \textit{stuff} classes (\eg, \textit{sky}, \textit{building}) in the inference prompts (third row) results in a slightly lower performance compared to using only \textit{things} classes (second row).

Given that the target dataset Cityscapes is captured with sensors mounted on a car, we make an attempt to tailor the inference prompts for such a use-case. Specifically, we use the inference prompt $p$ = ``\textit{a photo of a} $V_*$ \textit{seen from the dash cam}''. We notice that usage of such prompt does not bring any improvement, and rather leads to worsened performance.

Next, we make the training prompt more suited for a driving scenario by using $p$ = ``\textit{a photo of a} $V_*$ \textit{scene from a car}'' and experiment with some inference prompts that are essentially nuanced variations of the training prompt. 
The results are reported in the lower part of Tab.~\ref{tab:prompts}. We observe that adding the phrase ``\textit{scene from a car}'' to the training prompt has no positive impact in the training of \method.
It is worth noting that the retention of the \textit{prior preservation} loss caused our best result to decrease from 57.2\% to 54.8\%.

\input{figs/ablation_4}

\vspace{1mm}
\noindent\textbf{Impact of generated dataset cardinality.} 
Here, we examine the impact of the cardinality of the target dataset generated by \method using a single real target image (\ie, one-shot) on the performance of the segmentation model.
In Fig.~\ref{fig:size} we plot the \miou from \daf versus the generated dataset size and also compare with training on the real target dataset of the same cardinality. We observe that having the same quantity of real target samples leads to better performance with respect to purely synthetic data. This is expected as real data always contains more targetted information than synthetic data. However, one must appreciate the fact that having 1000 synthetic data leads to a better performance than having 10 real samples, which can be difficult to collect in some applications. Thus, our \method is most effective when working with a very small budget of real target data.

%% file: tables/g2c_sota.tex
\begin{table*}[!t]
\centering
\small
\caption{Comparison with state-of-the-art methods for \colorbox{\all}{UDA} and \colorbox{\one}{\osda} on the \gtoc benchmark. \#TS denotes the number of \textit{real} target samples used during training, which are color coded as \colorbox{\none}{None}, \colorbox{\all}{All} and \colorbox{\one}{One}. Methods using ResNet-101~\cite{he2016deep} and MiT-B5~\cite{xie2021segformer} are shown in the top and bottom halves, respectively. As an example, DaFormer + \method denotes DAFormer trained using the synthetic images generated by our \method. $\star$: results from CACDA~\cite{gong2022one}; $\diamond$: results from HRDA~\cite{hoyer2022hrda}; and $\dagger$: results from ASM~\cite{luo2020adversarial}.}
\vspace{-1mm}
\label{tab:g2c-sota}
\setlength{\tabcolsep}{3pt}
\resizebox{\textwidth}{!}{%
\begin{tabular}{l|c|ccccccccccccccccccc|c}
\toprule
 & \#TS & \rotatebox{90}{Road} & \rotatebox{90}{S.walk} & \rotatebox{90}{Build.} & \rotatebox{90}{Wall} & \rotatebox{90}{Fence} & \rotatebox{90}{Pole} & \rotatebox{90}{Tr.Light} & \rotatebox{90}{Sign} & \rotatebox{90}{Veget.} & \rotatebox{90}{Terrain} & \rotatebox{90}{Sky} & \rotatebox{90}{Person} & \rotatebox{90}{Rider} & \rotatebox{90}{Car} & \rotatebox{90}{Truck} & \rotatebox{90}{Bus} & \rotatebox{90}{Train} & \rotatebox{90}{M.bike} & \rotatebox{90}{Bike} & mIoU\\

\toprule
\multicolumn{22}{c}{\small Encoder: ResNet-101}\\
\midrule

\rowcolor{\none} Source-Only $\star$ & None & 75.8 & 16.8 & 77.2 & 12.5 & 21.0 & 25.5 & 30.1 & 20.1 & 81.3 & 24.6 & 70.3 & 53.8 & 26.4 & 49.9 & 17.2 & 25.9 & 6.5 & 25.3 & 36.0 & 36.6 \\

\rowcolor{\deeplab} CycleGAN $\dagger$~\cite{zhu2017unpaired} & All & 81.7 & 27.0 & 81.7 & 30.3 & 12.2 & 28.2 & 25.5 & 27.4 & 82.2 & 27.0 & 77.0 & 55.9 & 20.5 & 82.8 & 30.8 & 38.4 & 0.0 & 18.8 & 32.3 & 41.0 \\

\rowcolor{\deeplab} ASM $\dagger$~\cite{luo2020adversarial} & All & 89.8 & 38.2 & 77.8 & 25.5 & 28.6 & 24.9 & 31.2 & 24.5 & 83.1 & 36.0 & 82.3 & 55.7 & 28.0 & 84.5 & 45.9 & 44.7 & 5.3 & 26.4 & 31.3 & 45.5\\

\rowcolor{\deeplab} DACS $\diamond$~\cite{tranheden2021dacs} & All & 89.9 & 39.7 & 87.9 & 30.7 & 39.5 & 38.5 & 46.4 & 52.8 & 88.0 & 44.0 & 88.8 & 67.2 & 35.8 & 84.5 & 45.7 & 50.2 & 0.0 & 27.3 & 34.0 & 52.1\\

\rowcolor{\deeplab} ProDA $\diamond$~\cite{zhang2021prototypical} & All & 87.8 & 56.0 & 79.7 & 46.3 & 44.8 & 45.6 & 53.5 & 53.5 & 88.6 & 45.2 & 82.1 & 70.7 & 39.2 & 88.8 & 45.5 & 59.4 & 1.0 & 48.9 & 56.4 & 57.5\\
\rowcolor{\deeplab} DAFormer $\diamond$~\cite{hoyer2022daformer} & All & 96.0 & 72.4 & 88.0 & 39.2 & 37.4 & 38.0 & 50.3 & 54.0 & 88.4 & 47.2 & 89.2 & 69.8 & 42.6 & 88.6 & 48.6 & 55.4 & 0.9 & 34.6 & 48.7 & 57.3\\
\rowcolor{\deeplab} HRDA $\diamond$~\cite{hoyer2022hrda} & All & 96.2 & 73.1 & 89.7 & 43.2 & 39.9 & 47.5 & 60.0 & 60.0 & 89.9 & 47.1 & 90.2 & 75.9 & 49.0 & 91.8 & 61.9 & 59.3 & 10.2 & 47.0 & 65.3 & 63.0\\




\rowcolor{\one} AdaptSeg $\star$~\cite{tsai2018learning} & One & 77.7 & 19.2 & 75.5 & 11.7 & 6.4 & 16.8 & 18.2 & 15.4 & 77.1 & 34.0 & 68.5 & 55.3 & 30.9 & 74.5 & 23.7 & 28.3 & 2.9 & 14.4 & 18.9 & 35.2 \\
\rowcolor{\one} CBST $\star$~\cite{zou2018unsupervised} & One & 76.1 & 22.2 & 73.5 & 13.8 & 18.8 & 19.1 & 20.7 & 18.6 & 79.5 & 41.3 & 74.8 & 57.4 & 19.9 & 78.7 & 21.3 & 28.5 & 0.0 & 28.0 & 13.2 & 37.1 \\
\rowcolor{\one} CycleGAN $\star$~\cite{zhu2017unpaired} & One & 80.3 & 23.8 & 76.7 & 17.3 & 18.2 & 18.1 & 21.3 & 17.5 & 81.5 & 40.1 & 74.0 & 56.2 & 38.3 & 77.1 & 30.3 & 27.6 & 1.7 & 30.0 & 22.2 & 39.6 \\
\rowcolor{\one} OST $\star$~\cite{benaim2018one} & One & 84.3 & 27.6 & 80.9 & 24.1 & 23.4 & 26.7 & 23.2 & 19.4 & 80.2 & 42.0 & 80.7 & 59.2 & 20.3 & 84.1 & 35.1 & 39.6 & 1.0 & 29.1 & 23.2 & 42.3 \\
\rowcolor{\one} SMPPM $\star$~\cite{wu2022style} & One & 85.0 & 23.2 & 80.4 & 21.3 & 24.5 & 30.0 & 32.0 & 26.7 & 83.2 & 34.8 & 74.0 & 57.3 & 29.0 & 77.7 & 27.3 & 36.5 & 5.0 & 28.2 & 39.4 & 42.8 \\
\rowcolor{\one} ASM $\dagger$~\cite{luo2020adversarial} & One & 86.2 & 35.2 & 81.4 & 24.2 & 25.5 & 31.5 & 31.5 & 21.9 & 82.9 & 30.5 & 80.1 & 57.3 & 22.9 & 85.3 & 43.7 & 44.9 & 0.0 & 26.5 & 34.9 & 44.5 \\
\rowcolor{\one} DAFormer~\cite{hoyer2022daformer} & One & 85.5 & 31.2 & 81.7 & 24.0 & 25.6 & 23.0 & 33.1 & 27.4 & 82.7 & 27.8 & 81.4 & 61.6 & 27.2 & 79.0 & 30.5 & 41.4 & 13.4 & 29.2 & 14.9 & 43.2 \\
\rowcolor{\one} HRDA~\cite{hoyer2022hrda} & One & 86.7 & 22.0 & 81.2 & 26.8 & 25.8 & 30.2 & 40.4 & 33.6 & 84.8 & 24.3 & 77.8 & 63.2 & 32.3 & 84.7 & 31.1 & 40.6 & 19.4 & 26.5 & 14.0 & 44.5 \\
\rowcolor{\one} CACDA $\star$~\cite{gong2022one} & One & 80.9 & 32.6 & 85.8 & 36.1 & 30.7 & 40.7 & 43.7 & 41.7 & 84.1 & 30.7 & 84.5 & 65.4 & 27.6 & 86.0 & 36.5 & 51.4 & 24.1 & 26.7 & 30.7 & 49.5 \\
\midrule
\rowcolor{\one} \textbf{DAFormer + \method} & One & 88.1 & 32.8 & 84.3 & 26.6 & 27.7 & 32.7 & 35.7 & 34.9 & 86.2 &	36.2 & 87.6 & 65.8 & 35.8 &	80.2 & 39.5 & 	44.1 & 17.1 & 42.7 & 43.8 &	49.6 \\
\rowcolor{\one} \textbf{HRDA + \method} & One & 82.1 & 31.9 & 80.9 & 21.3 & 27.6 & 38.6 & 43.5 & 41.0 & 87.1 & 33.1 & 87.2 & 70.8 & 37.5 & 71.3 & 38.2 & 48.5 & 22.9 & 44.2 & 54.0 & 50.6 \\

\midrule
\midrule

\multicolumn{22}{c}{\small Encoder: MiT-B5}\\
\midrule

\rowcolor{\none} Source-Only $\diamond$ & None & - & -& & & & & & & & & & & & & & & & & & 45.6 \\
\rowcolor{\all} DAFormer $\diamond$~\cite{hoyer2022daformer} & All & 95.7 & 70.2 & 89.4 & 53.5 & 48.1 & 49.6 & 55.8 & 59.4 & 89.9 & 47.9 & 92.5 & 72.2 & 44.7 & 92.3 & 74.5 & 78.2 & 65.1 & 55.9 & 61.8 & 68.3\\

\rowcolor{\all} HRDA $\diamond$~\cite{hoyer2022hrda} & All & 96.4 & 74.4 & 91.0 & 61.6 & 51.5 & 57.1 & 63.9 & 69.3 & 91.3 & 48.4 & 94.2 & 79.0 & 52.9 & 93.9 & 84.1 & 85.7 & 75.9 & 63.9 & 67.5 & 73.8 \\

\rowcolor{\one} DAFormer~\cite{hoyer2022daformer} & One & 84.0 & 18.2 & 83.0 & 35.6 & 20.0 & 33.8 & 40.1 & 35.3 & 86.4 & 37.4 & 82.7 & 66.6 & 31.6 & 85.0 & 37.0 & 40.6 & 36.7 & 33.3 & 29.0 & 48.2 \\
\rowcolor{\one} HRDA~\cite{hoyer2022hrda} & One & 84.3 & 29.2 & 84.3 & 44.3 & 23.2 & 43.9 & 48.7 & 39.2 & 88.2 & 41.0 & 82.6 & 70.5 & 36.9 & 85.8 & 43.7 & 51.4 & 42.2 & 34.7 & 30.6 & 52.9 \\
\rowcolor{\one} CACDA $\star$~\cite{gong2022one} & One & 83.4 & 35.3 & 87.1 & 44.8 & 32.3 & 42.5 & 50.2 & 52.5 & 88.0 & 46.1 & 90.4 & 66.7 & 25.6 & 88.6 & 50.3 & 50.8 & 44.5 & 34.4 & 38.6 & 55.4 \\

\midrule

\rowcolor{\segfor} \textbf{DAFormer + \method }& One & 86.2 & 29.0 & 87.1 & 41.5 & 35.6 & 44.7 & 48.5 & 42.7 & 88.4 & 42.4 & 88.3 & 68.8 & 35.9 & 89.7 & 57.1 & 57.6 & 27.8 & 46.8 & 53.2 & 56.4 \\
\rowcolor{\segfor} \textbf{HRDA + \method} & One & 87.1 & 32.0 & 88.2 & 49.6 & 40.4 & 49.5 & 54.8 & 43.6 & 89.9 & 44.6 & 91.3 & 74.9 & 45.7 & 91.4 & 61.7 & 67.0 & 37.1 & 57.7 & 55.8 & 61.2  \\



\bottomrule
\end{tabular}%
}
\vspace{-1mm}
\end{table*}

%% file: tables/s2c_sota.tex
\begin{table*}[!t]
\centering
\small

\caption{Comparison with state-of-the-art methods for \colorbox{\all}{UDA} and \colorbox{\one}{\osda} on the \gtoc benchmark. \#TS denotes the number of \textit{real} target samples used during training, which are color coded as \colorbox{\none}{None}, \colorbox{\all}{All} and \colorbox{\one}{One}. Methods using ResNet-101~\cite{he2016deep} and MiT-B5~\cite{xie2021segformer} are shown in the top and bottom halves, respectively. As an example, DaFormer + \method denotes DAFormer trained using the synthetic images generated by our \method. mIoU\textsuperscript{13} and mIoU\textsuperscript{16} denote the mIoU computed using the 13 and 16 classes, respectively~\cite{tranheden2021dacs,vu2019advent,tsai2018learning}. $\star$: results from CACDA~\cite{gong2022one}; $\diamond$: results from HRDA~\cite{hoyer2022hrda}; and $\dagger$: results from ASM~\cite{luo2020adversarial}.}
\vspace{-1mm}
\label{tab:s2c-sota}
\setlength{\tabcolsep}{3pt}
\resizebox{0.95\textwidth}{!}{%
\begin{tabular}{l|c|cccccccccccccccc|cc}
\toprule
 & \#TS & \rotatebox{90}{Road} & \rotatebox{90}{S.walk} & \rotatebox{90}{Build.} & \rotatebox{90}{Wall} & \rotatebox{90}{Fence} & \rotatebox{90}{Pole} & \rotatebox{90}{Tr.Light} & \rotatebox{90}{Sign} & \rotatebox{90}{Veget.} & \rotatebox{90}{Sky} & \rotatebox{90}{Person} & \rotatebox{90}{Rider} & \rotatebox{90}{Car} & \rotatebox{90}{Bus} & \rotatebox{90}{M.bike} & \rotatebox{90}{Bike} & mIoU\textsuperscript{16} & mIoU\textsuperscript{13}\\
\toprule

\multicolumn{20}{c}{\small Encoder: ResNet-101}\\
\midrule

\rowcolor{\none} Source-Only $\star$ & None & 36.3 & 14.6 & 68.8 & 9.2 & 0.2 & 24.4 & 5.6 & 9.1 & 69.0 & 79.4 & 52.5 & 11.3 & 49.8 & 9.5 & 11.0 & 20.7  & 29.5 & 33.7\\
\rowcolor{\all} AdaptSeg $\dagger$~\cite{tsai2018learning} & All & 84.3 & 42.7 & 77.5 & - & - & - & 4.7 & 7.0 & 77.9 &  82.5 & 54.3 & 21.0 & 72.3 & 32.2 & 18.9 & 32.3 & - & 46.7  \\
\rowcolor{\all} CBST $\dagger$~\cite{zou2018unsupervised} & All & 53.6 & 23.7 & 75.0 & - & - & - & 23.5 & 26.3 & 84.8 & 74.7 & 67.2 & 17.5 & 84.5 & 28.4 & 15.2 & 55.8 & - & 48.4 \\
\rowcolor{\all} DACS $\diamond$~\cite{tranheden2021dacs} & All & 80.6 & 25.1 & 81.9 & 21.5 & 2.9 & 37.2 & 22.7 & 24.0 & 83.7 & 90.8 & 67.6 & 38.3 & 82.9 & 38.9 & 28.5 & 47.6 & 48.3 & 54.8 \\
\rowcolor{\all} ProDA $\diamond$~\cite{zhang2021prototypical} & All & 87.8 & 45.7 & 84.6 & 37.1 & 0.6 & 44.0 & 54.6 & 37.0 & 88.1 & 84.4 & 74.2 & 24.3 & 88.2 & 51.1 & 40.5 & 45.6 & 55.5 & 62.0  \\

\rowcolor{\all} DAFormer $\diamond$~\cite{hoyer2022daformer} & All & 71.5 & 30.4 & 85.4 & 26.2 & 3.4 & 40.9 & 45.9 & 52.3 & 84.3 & 81.4 & 69.7 & 42.7 & 86.9 & 52.5 & 49.3 & 59.4 & 55.1 & 61.9  \\

\rowcolor{\all} HRDA $\diamond$~\cite{hoyer2022hrda} & All & 85.8 & 47.3 & 87.3 & 27.3 & 1.4 & 50.5 & 57.8 & 61.0 & 87.4 & 89.1 & 76.2 & 48.5 & 87.3 & 49.3 & 55.0 & 68.2 & 61.2 & 69.2 \\

\rowcolor{\one} CBST $\dagger$~\cite{zou2018unsupervised} & One & 59.6 & 24.1 & 72.9 & - & - & - & 5.5 & 13.8 & 72.2 & 69.8 & 55.3 & 21.1 & 57.1 & 17.4 & 13.8 & 18.5 & -& 38.5  \\
\rowcolor{\one} AdaptSeg $\dagger$~\cite{tsai2018learning} & One & 64.1 & 25.6 & 75.3 & - & - & - & 4.7 & 2.7 & 77.0 & 70.0 & 52.2 & 20.6 & 51.3 & 22.4 & 19.9 & 22.3 & - & 39.1 \\
\rowcolor{\one} OST $\dagger$~\cite{benaim2018one} & One & 75.3 & 31.6 & 72.1 & - & - & - & 12.3 & 9.3 & 76.1 & 71.1 & 51.1 & 17.7 & 68.9 & 19.0 & 26.3 & 25.4 & - & 42.8 \\
\rowcolor{\one} ASM $\dagger$~\cite{luo2020adversarial} & One & 73.5 & 29.0 & 75.2 & - & - & - & 10.9 & 10.1 & 78.1 & 73.2 & 56.0 & 23.7 & 76.9 & 23.3 & 24.7 & 18.2 & - & 44.1  \\
\rowcolor{\one} SMPPM $\star$~\cite{wu2022style} & One & 79.3 & 35.3 & 75.9 & 5.6 & 16.6 & 29.8 & 25.4 & 22.7 & 79.9 & 76.8 & 54.6 & 23.5 & 60.2 & 23.9 & 21.2 & 36.6 & 41.4 & 47.3  \\
\rowcolor{\one} DAFormer~\cite{hoyer2022daformer} & One & 69.3 & 26.3 & 76.3 & 5.8 & 0.5 & 28.5 & 16.7 & 24.9 & 73.7 & 74.9 & 59.5 & 28.5 & 74.5 & 28.0 & 21.8 & 44.6 & 40.9 & 47.1 \\
\rowcolor{\one} HRDA~\cite{hoyer2022hrda} & One & 61.0 & 24.1 & 76.7 & 7.5 & 0.3 & 34.5 & 21.8 & 29.2 & 77.4 & 78.9 & 64.2 & 28.5 & 77.1 & 25.0 & 29.8 & 43.4 & 42.5 & 48.5 \\
\rowcolor{\one} CACDA $\star$~\cite{gong2022one} & One & 82.5 & 33.8 & 77.8 & 12.6 & 0.8 & 34.2 & 30.8 & 34.4 & 79.8 & 82.4 & 55.4 & 30.7 & 72.5 & 28.4 & 15.9 & 47.8 & 45.0 & 51.7  \\
\midrule
\rowcolor{\one} \textbf{DAFormer + \method} & One & 79.3 & 32.9 & 80.6 & 17.7 & 0.4 & 32.4 & 22.2 & 36.9 & 82.4 & 81.6 & 65.7 & 36.0 & 76.2 & 26.0 & 31.2 & 50.3 & 47.0 & 53.5  \\
\rowcolor{\one} \textbf{HRDA + \method} & One & 86.5 & 39.3 & 83.2 & 17.9 & 0.2 & 42.8 & 24.0 & 45.1 & 84.1 & 85.9 & 72.7 & 39.2 & 86.1 & 31.4 & 44.5 & 56.7 & 52.5 & 59.4  \\

\midrule
\midrule
\multicolumn{20}{c}{\small Encoder: MiT-B5}\\
\midrule

\rowcolor{\all} DAFormer $\diamond$~\cite{hoyer2022daformer} & All & 84.5 & 40.7 & 88.4 & 41.5 & 6.5 & 50.0 & 55.0 & 54.6 & 86.0 & 89.8 & 73.2 & 48.2 & 87.2 & 53.2 & 53.9 & 61.7 & 60.9 & 67.4 \\
\rowcolor{\all} HRDA $\diamond$~\cite{hoyer2022hrda} & All & 85.2 & 47.7 & 88.8 & 49.5 & 4.8 & 57.2 & 65.7 & 60.9 & 85.3 & 92.9 & 79.4 & 52.8 & 89.0 & 64.7 & 63.9 & 64.9 & 65.8 & 72.4  \\

\rowcolor{\segfor} DAFormer~\cite{hoyer2022daformer} & One & 71.2 & 25.7 & 82.3 & 20.5 & 0.9  & 37.0 & 30.0 & 28.5 & 83.7 & 86.8 & 61.2 & 31.0 & 73.3 & 24.8 & 14.1 & 28.8 & 43.7 & 48.9 \\
\rowcolor{\segfor} HRDA~\cite{hoyer2022hrda} & One & 73.2 & 27.6 & 81.8 & 24.0 & 0.5 & 43.5 & 42.0 & 32.5 & 85.3 & 87.2 & 65.3 & 30.3 & 74.5 & 29.8 & 13.4 & 42.6 & 47.1 & 52.3  \\
\rowcolor{\segfor} CACDA $\star$~\cite{gong2022one} & One & 81.4 & 37.3 & 84.8 & 19.5 & 1.2 & 43.7 & 43.0 & 34.4 & 86.5 & 90.0 & 63.8 & 32.8 & 79.6 & 42.7 & 28.0 & 47.2 & 51.0 & 57.8\\

\midrule

\rowcolor{\segfor} \textbf{DAFormer + \method} & One & 79.6 & 28.8 & 85.6 & 30.9 & 1.4 & 45.6 & 43.0 & 46.5 & 85.9 & 89.7 & 70.3 & 38.4 & 84.8 & 56.0 & 39.5 & 52.1 & 54.9 & 61.1  \\
\rowcolor{\segfor} \textbf{HRDA + \method} & One & 83.2 & 31.8 & 86.6 & 37.4 & 0.8 & 51.4 & 46.9 & 52.0 & 87.8 & 92.0 & 76.1 & 43.7 & 88.4 & 56.3 & 48.5 & 57.1 & 58.7 & 64.9 \\
\bottomrule
\end{tabular}%
}
\vspace{-1mm}
\end{table*}

%% file: tables/ablation/style_transfer.tex
\begin{table}[!h]
    \centering
    \small
    \caption{Comparison with style transfer-based \osda methods}
    \vspace{-2mm}
    \begin{tabular}{l|cc}
        \toprule
        & ResNet-101 & MiT-B5 \\ 
        \midrule
        RAIN~\cite{luo2020adversarial} & 42.7 & 53.4 \\ 
        PT+CDIR~\cite{gong2022one} & 48.5 & 54.0 \\
        \method (Ours) & \textbf{50.6} & \textbf{57.2} \\ 
        \bottomrule
    \end{tabular}
    \label{tab:style_transfer}
    \vspace{-3mm}
\end{table}

%% file: figs/ablation_1_2.tex
\begin{figure}[tp]
    \centering
    \input{figs/assets/training_steps_vs_miou}    
    \vspace{-3mm}
    \caption{Impact of number of shots (\#TS) on the mIoU (in \%)}
    \label{fig:num-shots-miou}
\end{figure}
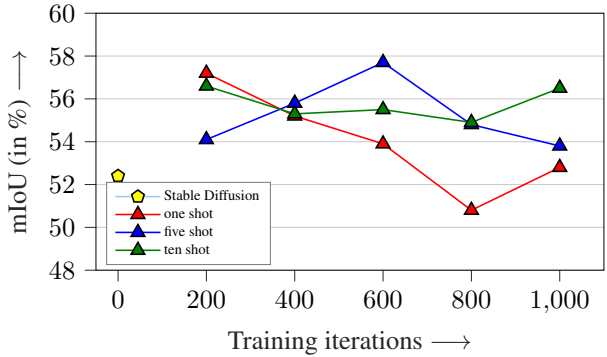

%% file: figs/assets/training_steps_vs_miou.tex
\begin{tikzpicture}

\definecolor{darkslategray38}{RGB}{38,38,38}
\definecolor{green}{RGB}{0,128,0}
\definecolor{lightblue161201244}{RGB}{161,201,244}
\definecolor{lightgray204}{RGB}{204,204,204}
\definecolor{yellow}{RGB}{255,255,0}

\begin{axis}[
axis line style={line width=.1mm},
legend cell align={left},
legend style={
  fill opacity=1,
  draw opacity=1,
  text opacity=1,
  at={(0.02,0.02)},
  anchor=south west,
  draw=gray,
  nodes={scale=0.55, transform shape}
},
height=0.6\linewidth,
width=\linewidth,
tick align=outside,
x grid style={lightgray204},
xlabel=\textcolor{darkslategray38}{Training iterations \(\displaystyle \longrightarrow\)},
xmajorticks=true,
xmin=-50, xmax=1100,
xtick pos=left,
xtick style={color=darkslategray38},
y grid style={lightgray204},
ylabel=\textcolor{darkslategray38}{mIoU (in \%) \(\displaystyle \longrightarrow\)},
ymajorgrids,
ymajorticks=true,
ymin=48, ymax=60,
ytick pos =left,
ytick = {48, 50, 52, 54, 56, 58, 60},
ytick style={color=darkslategray38}
]
\addplot [semithick, lightblue161201244, mark=pentagon*, mark size=2.5, mark options={solid,fill=yellow,draw=black}]
table {%
0 52.4
};
\addlegendentry{Stable Diffusion}
\addplot [semithick, red, mark=triangle*, mark size=3, mark options={solid,draw=black}]
table {%
200 57.2
400 55.2
600 53.9
800 50.8
1000 52.8
};
\addlegendentry{one shot}
\addplot [semithick, blue, mark=triangle*, mark size=3, mark options={solid,draw=black}]
table {%
200 54.1
400 55.8
600 57.7
800 54.8
1000 53.8
};
\addlegendentry{five shot}
\addplot [semithick, green, mark=triangle*, mark size=3, mark options={solid,draw=black}]
table {%
200 56.6
400 55.3
600 55.5
800 54.9
1000 56.5
};
\addlegendentry{ten shot}
\end{axis}

\end{tikzpicture}

%% file: tables/ablation/prompts.tex
\begin{table}[!t]\centering
\caption{Impact of training and inference prompts on the \miou.
}
\label{tab:prompts}
\resizebox{\columnwidth}{!}{%
\begin{tabular}{ccc|c}
\toprule

\textbf{Training prompt} &\textbf{Inference prompt} &\textbf{classes} &\textbf{mIoU} \\
\midrule

\multirow{4}{*}{\makecell{``a photo of a $V_*$\\ urban scene''}} & ``a photo of a $V_*$ urban scene'' & - & 52.9 \\
\cmidrule{2-4}

& ``a photo of a $V_*$ [CLS]'' & things & \textbf{57.2} \\
\cmidrule{2-4}

& ``a photo of a $V_*$ [CLS]'' & things + stuff & 56.7 \\
\cmidrule{2-4}
& ``a photo of a $V_*$ [CLS] seen from the dash cam'' & things & 55.5 \\

\midrule
\midrule

\multirow{3}{*}{\makecell{``a photo of a $V_*$\\ scene from a car''}} & \makecell{``a photo of $V_*$ scene from a car''} & things & 53.0 \\


\cmidrule{2-4}

& ``a photo of a $V_*$ [CLS]'' & things & 56.8 \\
\cmidrule{2-4}

& \makecell{``a photo of [CLS] \\in a $V_*$ scene from a car''} & things & 55.4 \\
\bottomrule
\end{tabular}%
}
\vspace{-5mm}
\end{table}

%% file: figs/ablation_4.tex
\begin{figure}[tp]
    \centering
    \input{figs/assets/dataset_size}
    \vspace{-3mm}
    \caption{Impact of the cardinality of the generated target dataset on the \miou in the one-shot setting. It is compared with adaptation on the real data}
    \vspace{-4mm}
    \label{fig:size}
\end{figure}
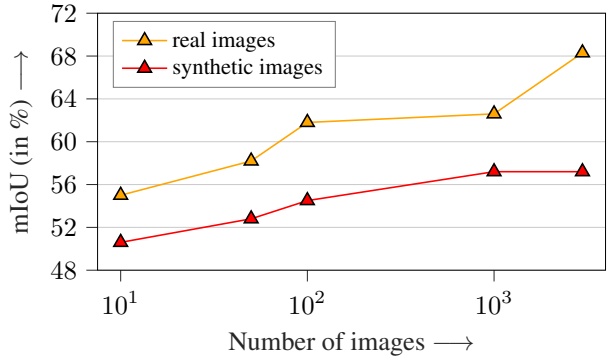

%% file: figs/assets/dataset_size.tex
\begin{tikzpicture}

\definecolor{darkslategray38}{RGB}{38,38,38}
\definecolor{lightgray204}{RGB}{204,204,204}
\definecolor{orange}{RGB}{255,165,0}

\begin{axis}[
axis line style={line width=.1mm},
legend cell align={left},
legend style={
  fill opacity=1,
  draw opacity=1,
  text opacity=1,
  at={(0.03,0.97)},
  anchor=north west,
  draw=gray,
  nodes={scale=0.85, transform shape}
},
height=0.6\linewidth,
width=\linewidth,
log basis x={10},
tick align=outside,
xtick pos=left,
x grid style={lightgray204},
xlabel=\textcolor{darkslategray38}{Number of images \(\displaystyle \longrightarrow\)},
xmajorticks=true,
xmin=7.52186702497734, xmax=3955.13506170891,
xmode=log,
xtick style={color=darkslategray38},
xtick={0.1,1,10,100,1000,10000,100000},
xticklabels={
  \(\displaystyle {10^{-1}}\),
  \(\displaystyle {10^{0}}\),
  \(\displaystyle {10^{1}}\),
  \(\displaystyle {10^{2}}\),
  \(\displaystyle {10^{3}}\),
  \(\displaystyle {10^{4}}\),
  \(\displaystyle {10^{5}}\)
},
y grid style={lightgray204},
ylabel=\textcolor{darkslategray38}{mIoU (in \%) \(\displaystyle \longrightarrow\)},
ymajorgrids,
ytick pos=left,
ymajorticks=true,
ymin=48, ymax=72,
ytick style={color=darkslategray38},
ytick = {48, 52, 56, 60, 64, 68, 72},
]
\addplot [semithick, orange, mark=triangle*, mark size=3, mark options={solid,draw=black}]
table {%
10 55.0
50 58.2
100 61.8
1000 62.6
2975 68.3
};
\addlegendentry{real images}
\addplot [semithick, red, mark=triangle*, mark size=3, mark options={solid,draw=black}]
table {%
10 50.6
50 52.8
100 54.5
1000 57.2
2975 57.2
};
\addlegendentry{synthetic images}
\end{axis}

\end{tikzpicture}

%% file: 10_conclusion.tex
\section{Conclusions}
\label{sec:conclusion}

We proposed a synthetic data generation method \method for the task of one-shot unsupervised domain adaptation, that uses a single image from the target domain to personalize a pre-trained text-to-image diffusion model. The personalization leads to a synthetic target dataset that faithfully depicts the style and content of the target domain, whereas the text-conditioning ability allows for generating diverse scenes with desired semantic objects. 
When pairing \method with modern UDA methods, it outperforms all state-of-the-art \osda methods, thus paving the path for future research in this few-shot learning paradigm.

%% file: 12_appendix.tex
\appendix
\label{sec:appendix}

\setcounter{table}{0}
\renewcommand{\thetable}{A\arabic{table}}%
\setcounter{figure}{0}
\renewcommand{\thefigure}{A\arabic{figure}}%
\setcounter{equation}{0}
\renewcommand{\theequation}{A\arabic{equation}}%

\twocolumn[\vspace*{2em}\centering\Large\bf%
{\Large Supplementary Material}%
\vspace*{2em}]
The supplementary material is organized as follows: Sec.~\ref{sec:add-exp} reports additional experiments and ablation analysis of our proposed method. Sec.~\ref{sec:oth-imp} provides additional implementation details. Sec.~\ref{sec:quali-seg} presents the segmentations maps and then we conclude with a discussion about the broader impact of our work.

\section{Additional experiments}
\label{sec:add-exp}

\begin{figure}[!h]
    \centering
    \input{figs/assets/training_steps_vs_fid}
    \caption{Impact of number of shots (\#TS) on the FID score}
    \label{fig:num-shots}
\end{figure}
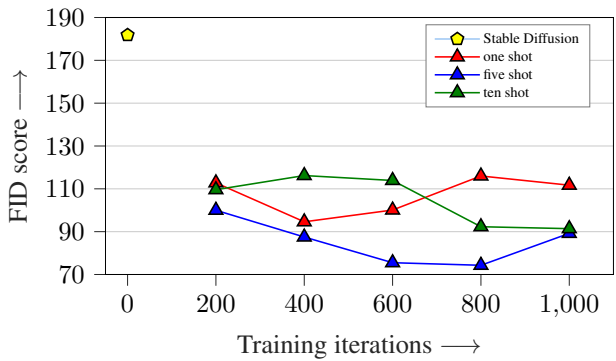

\noindent\textbf{Impact of number of shots on FID.} We also explore the connection between the number of shots (\ts) and the photo-realism of the generated target images using the Fréchet Inception Distance (FID)~\cite{heusel2017gans} score. The FID score measures how close are the generated images to the real target data distribution. Lower the FID score, closer are the two distributions. We plot the FID scores in Fig.~\ref{fig:num-shots}, and we observe that Stable Diffusion (SD) has very high FID score, showing that the generated images have very little resemblance to the target domain Cityscapes. Low similarity with the target domain is also reflected in poorer performance, as shown in Fig. 4 of the main paper.

When compared with SD, the generations from \method are much closer to the real target domain, which is evident from the lower FID scores. We notice that when we fine-tune SD with fewer real target images, the FID score shows an upward trend as the number of training iterations increases. Whereas, as the \ts increases from 1 to 5, longer training leads to decreased FID score, up until the 800\textsuperscript{th} interations. Finally, for the 10-shot setting, the FID score plateaus for a while and then starts going down after the 600\textsuperscript{th} interations. All these observations are as per expectations, since having more real images necessitates longer training to fit to that data distribution.

\begin{figure}
    \centering
    \setlength{\tabcolsep}{0.9pt}
    \resizebox{\columnwidth}{!}{%
    \begin{tabular}{lcc}
        \rotatebox{90}{\hspace{0.6cm} \small GTA} & \includegraphics[width=0.45\columnwidth]{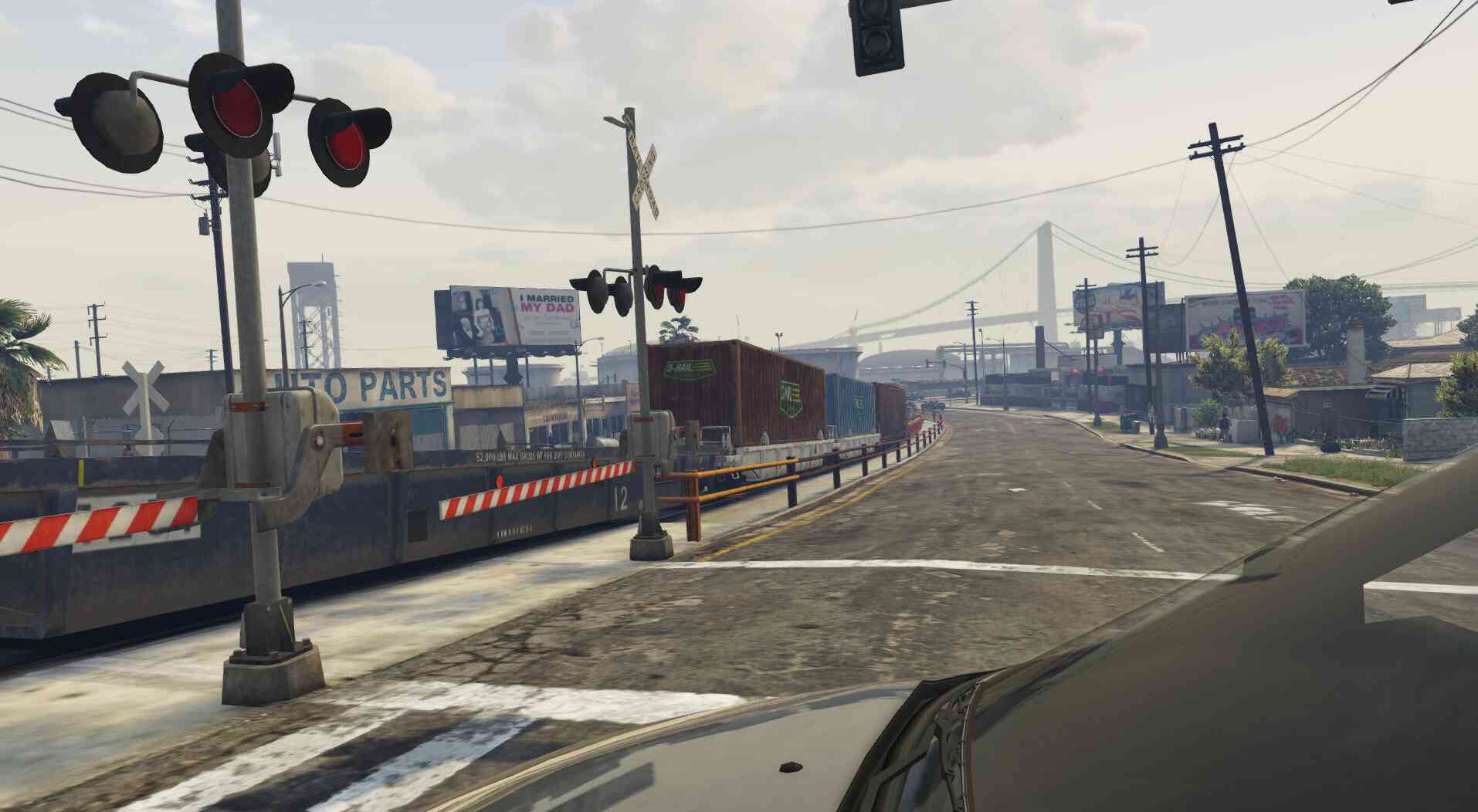} & \includegraphics[width=0.45\columnwidth]{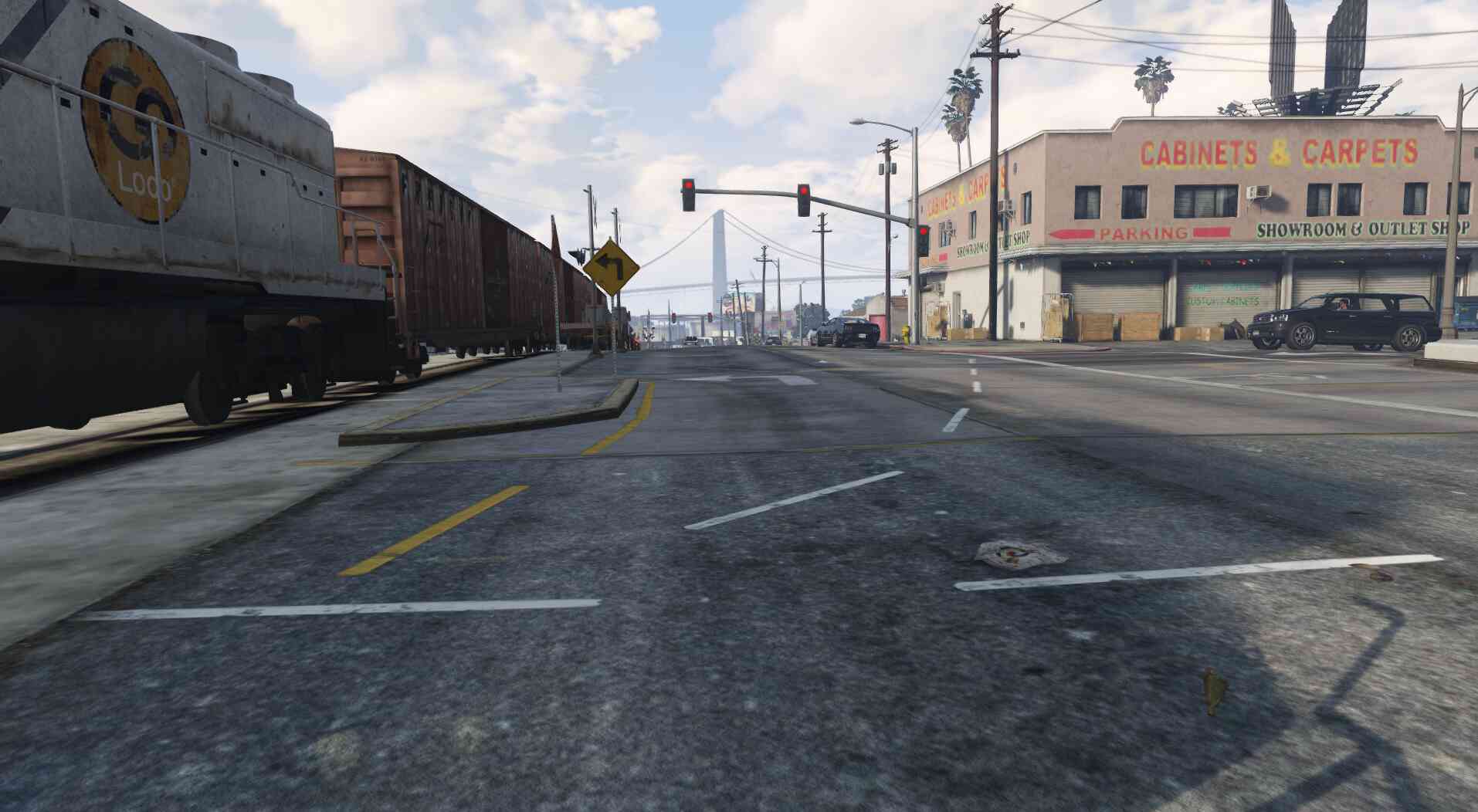} \\
        
        \rotatebox{90}{\hspace{0.1cm} \small Cityscapes} & \includegraphics[width=0.45\columnwidth]{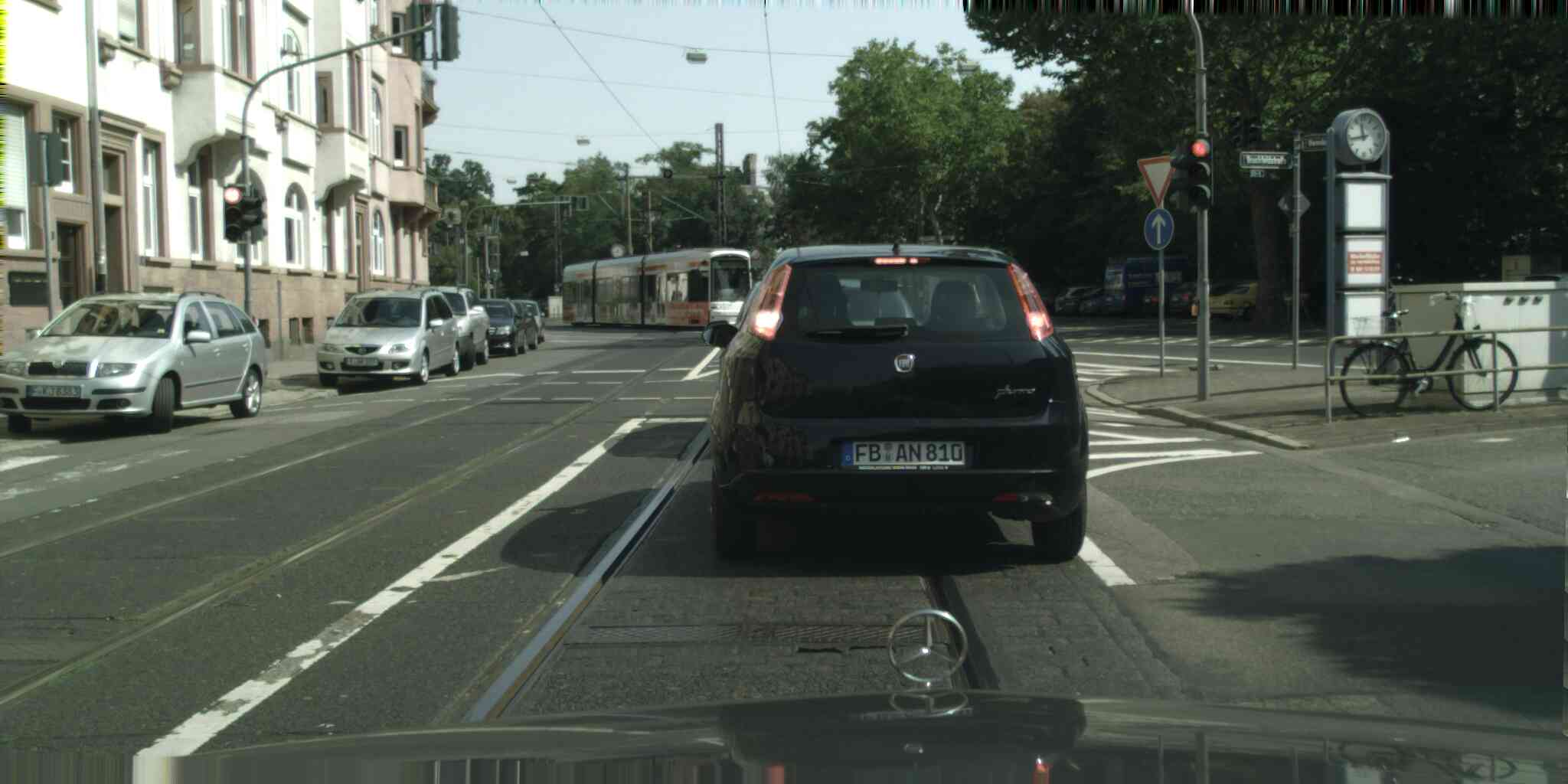} & \includegraphics[width=0.45\columnwidth]{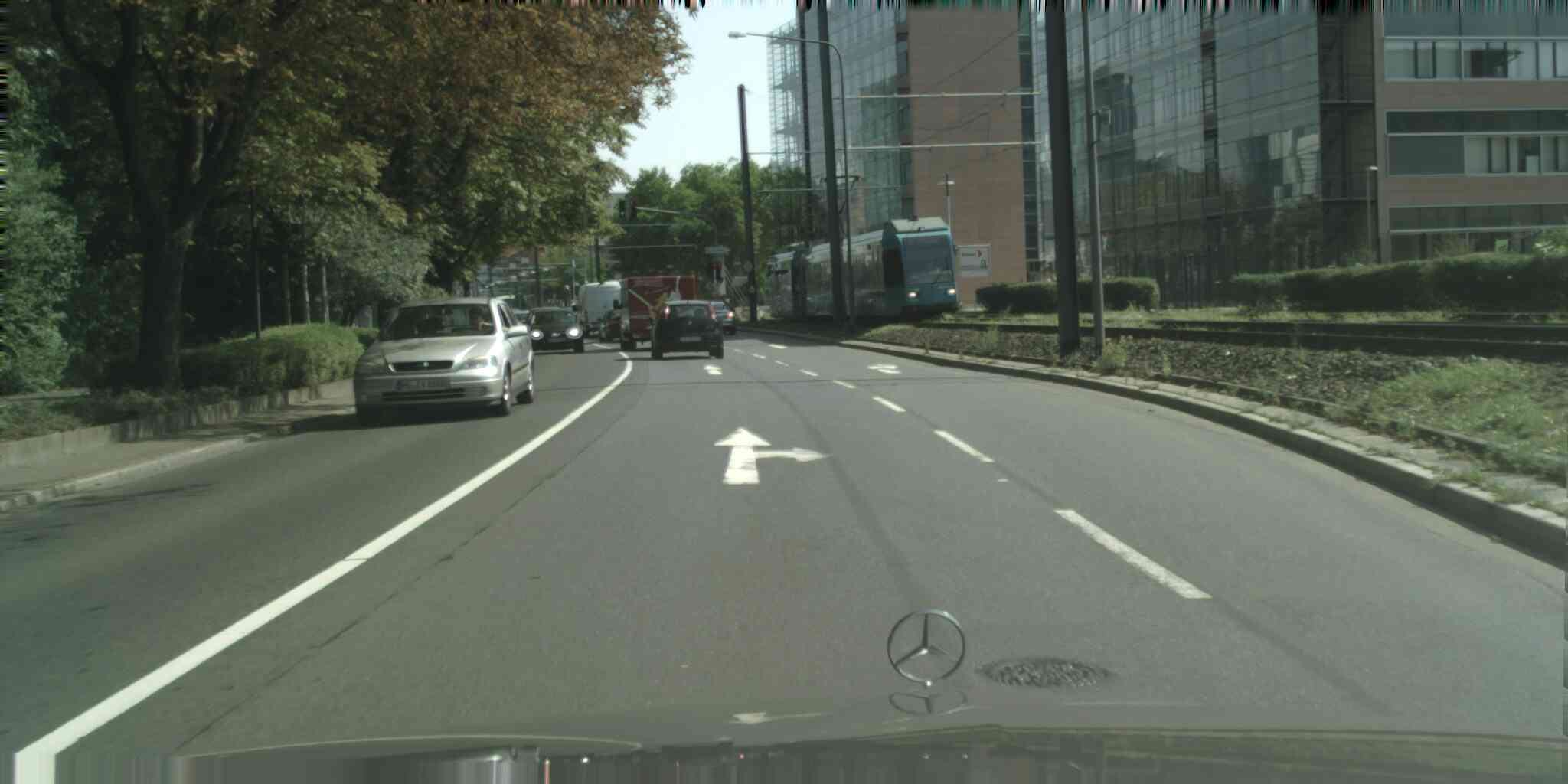} \\ 

        \rotatebox{90}{\hspace{.1cm} \small \method} &

        \begingroup  \sbox0{\includegraphics{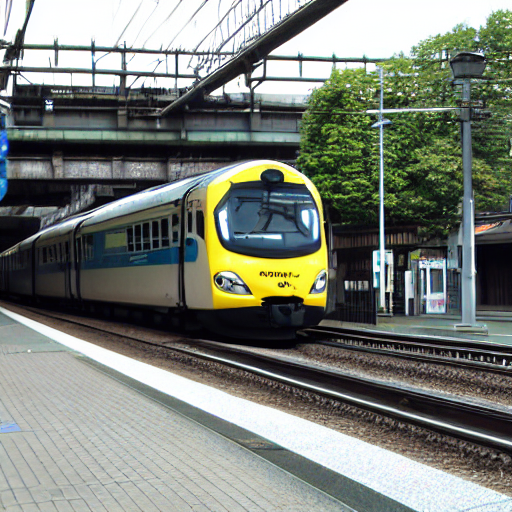}}%
        \includegraphics[clip,trim=0 {.27\wd0} 0 {.25\wd0},width=0.45\columnwidth]{figs/assets/train/datum1.png}
        \endgroup

        &

        \begingroup  \sbox0{\includegraphics{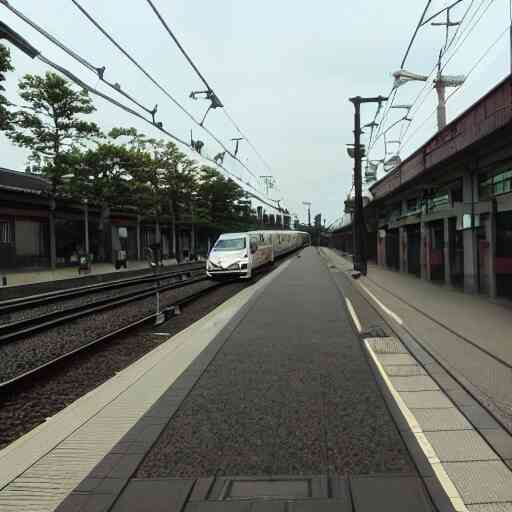}}%
        \includegraphics[clip,trim=0 {.27\wd0} 0 {.25\wd0},width=0.45\columnwidth]{figs/assets/train/datum2.jpg}
        \endgroup
        \\
    \end{tabular}%
    }
    \caption{Real and synthetic images from the things class \textit{train}}
    \label{fig:train}
\end{figure}

\noindent\textbf{Impact of prompting on class-wise IoU.} Next we examine the impact of using \textit{things} and \textit{stuff} classes on the class-wise IoU scores. We report the results computed using \daf~\cite{hoyer2022daformer} on the \gtoc benchmark in Tab.~\ref{tab:classwise-heatmaps}. We consider the \daf trained on a single real target image as the baseline, and the gain/loss attained by all the other methods are color coded. Warmer colors indicate gain, while cooler ones signify drops in performance. We compare the following methods: SD (using things class names during inference), \method (without things and stuff class names at inference), \method (using things and stuff class names at inference), \method (using things class names at inference, and w/ prior-preservation loss~\cite{ruiz2022dreambooth}), and \method (using things class names at inference, and w/o prior-preservation loss), which is our final method.

We observe from Tab.~\ref{tab:classwise-heatmaps} that using synthetic data, either with SD or our method brings improvements in a majority of the classes. Big improvements are noteworthy in the \textit{things} classes (shown in the left half of Tab.~\ref{tab:classwise-heatmaps}). Interestingly, for some things classes, such as \textit{person}, \textit{rider} and \textit{car}, the improvement with synthetic data is meagre. It could be potentially due to the fact that the source domain already encodes a strong prior about these objects, and additional data do not provide useful information.

Careful scrutiny of the table also reveals that there is a drop in the performance of the things class \textit{train}. In an attempt to investigate this drop, we visualize in Fig.~\ref{fig:train} the images annotated as \textit{train} in GTA and Cityscapes, as well as synthetic images of \textit{train} generated by \method. We observe an ambiguity in annotations for the \textit{train} class in GTA and Cityscapes. While in GTA, the train image really corresponds to the vehicle of type ``train'', in Cityscapes one can reasonably recognize that the vehicle is actually a \textit{tram}. Since, we utilize the class names of the source domain, our \method generates images with an object, \ie, \textit{train}, which is irrelevant to the target domain, despite both the vehicles exhibiting similar appearance.

\input{tables/g2c_classwise_heatmaps}
\input{figs/segmaps_qualitative}
\vspace{-1mm}
\section{Other Implementation details :}
\label{sec:oth-imp}
\vspace{-1mm}
\noindent\textbf{Data Augmentation.}
To enhance the robustness of the learned features and allow fair comparison, we adopt the identical set of data augmentation techniques as those employed in DAFormer~\cite{hoyer2022daformer}.
The augmentation process entails applying a Random Crop of size 512 $\times$ 512 to both source and target images, followed by Random Flip with a 0.5 probability. Next, we employ the photometric distortion utilized in DACS~\cite{tranheden2021dacs}, which comprises of a Gaussian Blur, Color Jittering, and ClassMix~\cite{olsson2021classmix}.

\noindent\textbf{Personalization and generation.}
In the personalization stage, we employ the default DDPM~\cite{ho2020denoising} noise scheduler as in Dreambooth~\cite{ruiz2022dreambooth}. In the data generation stage, we also use the default parameters of Dreambooth~\cite{ruiz2022dreambooth}: 50 inference steps and a guidance scale of 7.5.
\vspace{-1mm}
\section{Qualitative visualizations}
\label{sec:quali-seg}
Finally, we show the qualitative results of the segmentation maps generated by our method and the comparison with other state-of-the-art methods in Fig.~\ref{tab:segmaps}. Despite being trained on synthetic data, our \method is still able to capture several fine-grained details, especially the objects that appear far away from the camera. Note that, we do \textit{not} make efforts to cherry pick the segmentation maps, and simply report our results for the same RGB input maps, as reported in CACDA~\cite{gong2022one}.
\vspace{-1mm}
\section*{Broader Impact}
\vspace{-1mm}
Although SD is adept at generating high-fidelity images of geometrically coherent scenes, sometimes the generations are gibberish and defy commonsense reasoning. As shown in Fig.3(d) of the main paper, the fine-tuned SD generates a very convincing-looking yet unintelligible ``traffic sign'', which has no meaning in a driving manual. Thus, to avoid model poisoning~\cite{cao2022mpaf}, the practitioners should exercise utmost caution when deploying segmentation models, for autonomous driving, that are trained using such synthetic datasets.

%% file: figs/assets/training_steps_vs_fid.tex
\begin{tikzpicture}

\definecolor{darkslategray38}{RGB}{38,38,38}
\definecolor{green}{RGB}{0,128,0}
\definecolor{lightblue161201244}{RGB}{161,201,244}
\definecolor{lightgray204}{RGB}{204,204,204}
\definecolor{yellow}{RGB}{255,255,0}

\begin{axis}[
axis line style={line width=.1mm},
legend cell align={left},
legend style={
  fill opacity=1,
  draw opacity=1,
  text opacity=1,
  at={(0.63,0.65)},
  anchor=south west,
  draw=gray,
  nodes={scale=0.55, transform shape}
},
height=0.6\linewidth,
width=\linewidth,
tick align=outside,
x grid style={lightgray204},
xlabel=\textcolor{darkslategray38}{Training iterations \(\displaystyle \longrightarrow\)},
xmajorticks=true,
xmin=-50, xmax=1100,
xtick pos=left,
xtick style={color=darkslategray38},
y grid style={lightgray204},
ylabel=\textcolor{darkslategray38}{FID score \(\displaystyle \longrightarrow\)},
ymajorgrids,
ymajorticks=true,
ymin=70, ymax=190,
ytick pos =left,
ytick = {70, 90, 110, 130, 150, 170, 190},
ytick style={color=darkslategray38}
]
\addplot [semithick, lightblue161201244, mark=pentagon*, mark size=2.5, mark options={solid,fill=yellow,draw=black}]
table {%
0 181.8
};
\addlegendentry{Stable Diffusion}
\addplot [semithick, red, mark=triangle*, mark size=3, mark options={solid,draw=black}]
table {%
200 112.8
400 94.6
600 100.1
800 116.0
1000 111.7
};
\addlegendentry{one shot}
\addplot [semithick, blue, mark=triangle*, mark size=3, mark options={solid,draw=black}]
table {%
200 100.0
400 87.5
600 75.5
800 74.3
1000 89.2
};
\addlegendentry{five shot}
\addplot [semithick, green, mark=triangle*, mark size=3, mark options={solid,draw=black}]
table {%
200 109.6
400 116.2
600 113.9
800 92.3
1000 91.4
};
\addlegendentry{ten shot}
\end{axis}

\end{tikzpicture}

%% file: tables/g2c_classwise_heatmaps.tex
\begin{table*}[!t]
    \centering
    \small
    \resizebox{\textwidth}{!}{%
    \begin{tabular}{rcccccccccc|ccccccccc}

    & \rotatebox{90}{Tr.Light} & \rotatebox{90}{Sign} & \rotatebox{90}{Person} & \rotatebox{90}{Rider} & \rotatebox{90}{Car} & \rotatebox{90}{Truck} & \rotatebox{90}{Bus} & \rotatebox{90}{Train} & \rotatebox{90}{M.bike} & \rotatebox{90}{Bike} & \rotatebox{90}{\plus{0}Road} & \rotatebox{90}{S.walk} & \rotatebox{90}{Build.} & \rotatebox{90}{Wall} & \rotatebox{90}{Fence} & \rotatebox{90}{Pole} & \rotatebox{90}{Veget.} & \rotatebox{90}{Terrain} & \rotatebox{90}{Sky} \\
    
    
    Real target & 41.2 & 36.4 &	68.0 & 35.3 & 84.0 & 33.8 &	36.9 & 34.6 & 30.7 & 25.7 &	\plus{0}82.7 & 14.7 & 83.8 & 34.1 &	19.8 & 31.8 & 86.0 & 30.9 &	83.5 \\
    
    
    SD (things) & \plus{4.5}45.7 & \minus{8.6}27.8 & 68.0 & \plus{0.1}36.4 & \plus{4.5}88.5 & \plus{15}48.8 & \plus{17.2}54.1 & \minus{14.2}20.4 & \plus{13.6}44.3 & \plus{16.1}41.8 & \minus{4.3}78.4 & \plus{9.3}24.0 & \plus{1.4}85.2 & \plus{10.8}44.9 & \plus{14.2}34.0 & \plus{8.7}40.5 & \plus{2.2}88.2 & \plus{8.2}39.1 & \plus{2.9}86.4 \\
    
    
    \method & \plus{2.6}43.8 & \plus{11.0}47.4 & \minus{0.2}67.8 & \plus{0.9}36.2 & \plus{3.7}87.7 & \plus{13.2}47.0 & \plus{9.3}46.2 & \plus{7.6}42.2 & \plus{6.7}37.4 & \plus{5.4}31.1 & \plus{3.9}86.6 &	\plus{13.6}28.3 & \plus{1.2}85.0 & \plus{4.6}38.7 & \plus{2.4}22.2 & \plus{12.9}44.7 & \plus{1.6}87.6 & \plus{9.4}40.3 & \plus{1.6}85.1 \\
    
    
    \makecell[r]{\method \\(things \& stuff)} & \plus{7.1}48.3 & \plus{7.6}44.0 & \plus{0.6}68.6 & \plus{3.1}38.4 & \plus{6.2}90.2 & \plus{21.2}55.0 &	\plus{26.9}63.8 & \minus{11.3}23.3 & \plus{16}46.7 & \plus{29.3}55.0 &	\plus{3.2}85.9 & \plus{15}29.7 & \plus{3.3}87.1 & \plus{4.1}38.2 & \plus{20.2}40.0 & \plus{12.6}44.4 & \plus{1.7}88.7 & \plus{11.6}42.5 & \plus{3.4}86.9 \\
    

    \makecell[r]{\method (things)\\(w/ prior-loss)} & \plus{6.9}48.1 & \plus{9.8}46.4 & \minus{0.1}67.9 & \plus{2.3}37.6 & \plus{3.2}87.2 &  \plus{18.5}52.3 & \plus{13.5}50.4 & \minus{7.2}27.4 & \plus{17.6}48.3 &	\plus{23.1}48.8 &	\plus{3.7}86.4 & \plus{7.3}22.0 & \plus{2.3}86.1 & \plus{8.4}42.5 &	\plus{5.8}25.6 &	\plus{14.1}45.9 & \plus{2.4}88.4 &	\plus{10.0}41.9 & \plus{4.1}87.6\\
    
    \makecell[r]{\method (things)\\ (w/o prior loss)} & \plus{6.4}47.6 & \plus{6.4}42.8 & \plus{1.3}69.3 & \plus{0.9}36.2 & \plus{6.0}90.0 &	\plus{19.9}53.7 & \plus{22.9}59.8 & \minus{8.1}26.5 & \plus{20.1}50.8 & \plus{30.2}55.9 & \plus{4.7}87.4 & \plus{19.3}34.0 & \plus{3.4}87.2 & \plus{9.2}43.3 & \plus{18.5}38.5 & \plus{13.1}44.9 & \plus{0.6}88.6 &	\plus{12.7}43.6 & \plus{3.5}87.0 \\
    
    \end{tabular}%
    }
    \caption{Class-wise mIoU comparison for \gtoc using MiT-B5 encoder. The left part of the table indicates th \textit{things} classes, whereas the right part of the table indicates \textit{stuff} classes. The color visualizes the IoU difference with respect to the first row, which is trained with the single target image.}
    \label{tab:classwise-heatmaps}
\end{table*}

%% file: figs/segmaps_qualitative.tex
\begin{figure*}[tp]
    \centering
    \setlength{\tabcolsep}{0.5pt}
    \renewcommand{\arraystretch}{0.1}
    \resizebox{\textwidth}{!}{%
    \begin{tabular}{cccccc}

        \includegraphics[width=0.15\textwidth]{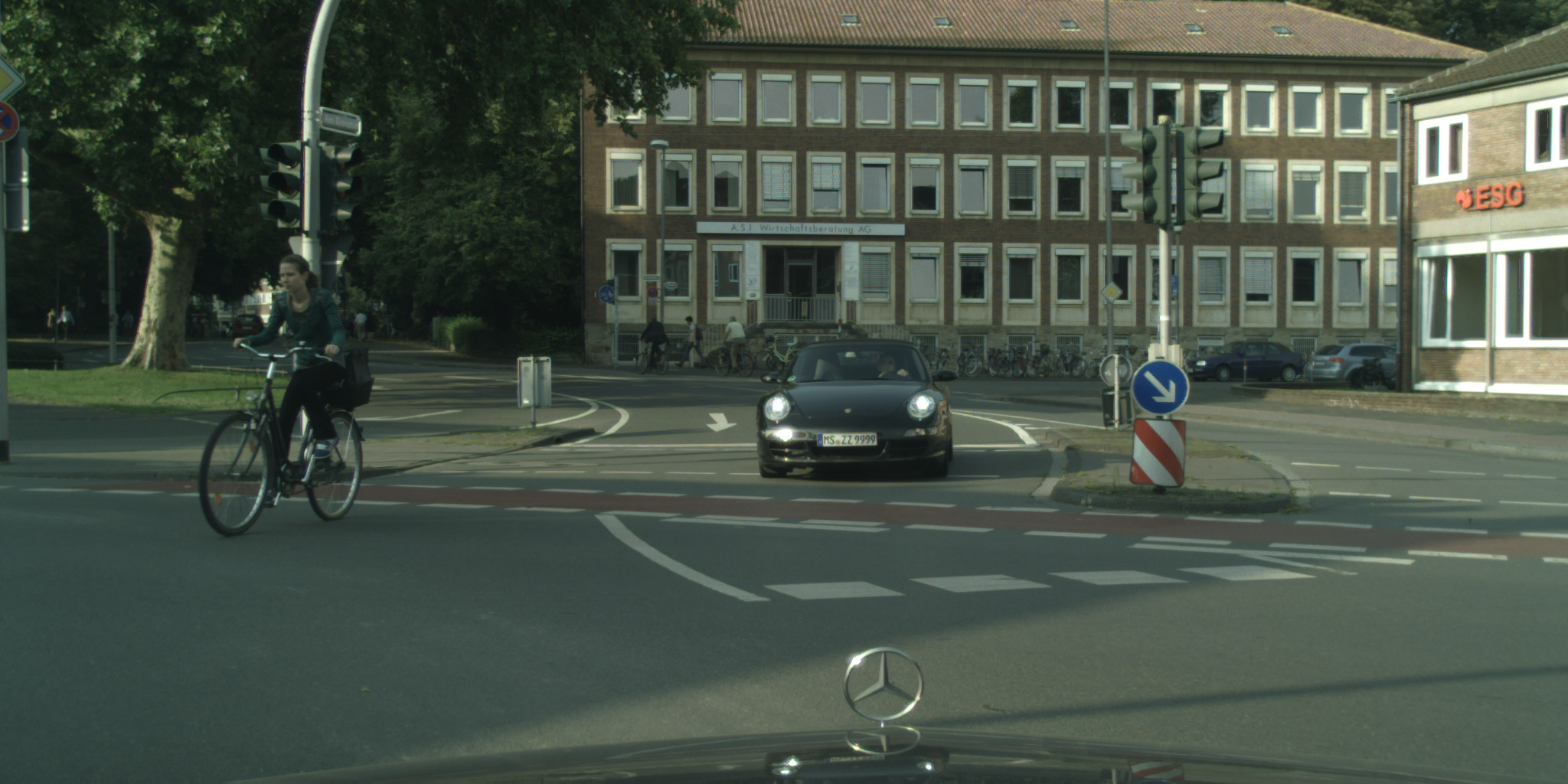} & \includegraphics[width=0.15\textwidth]{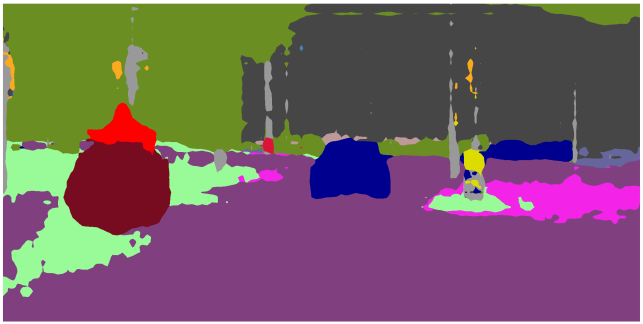} & \includegraphics[width=0.15\textwidth]{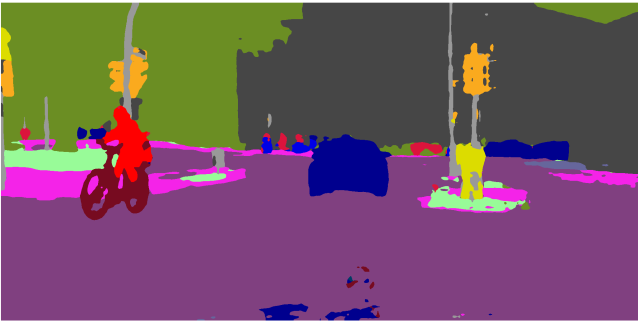} & \includegraphics[width=0.15\textwidth]{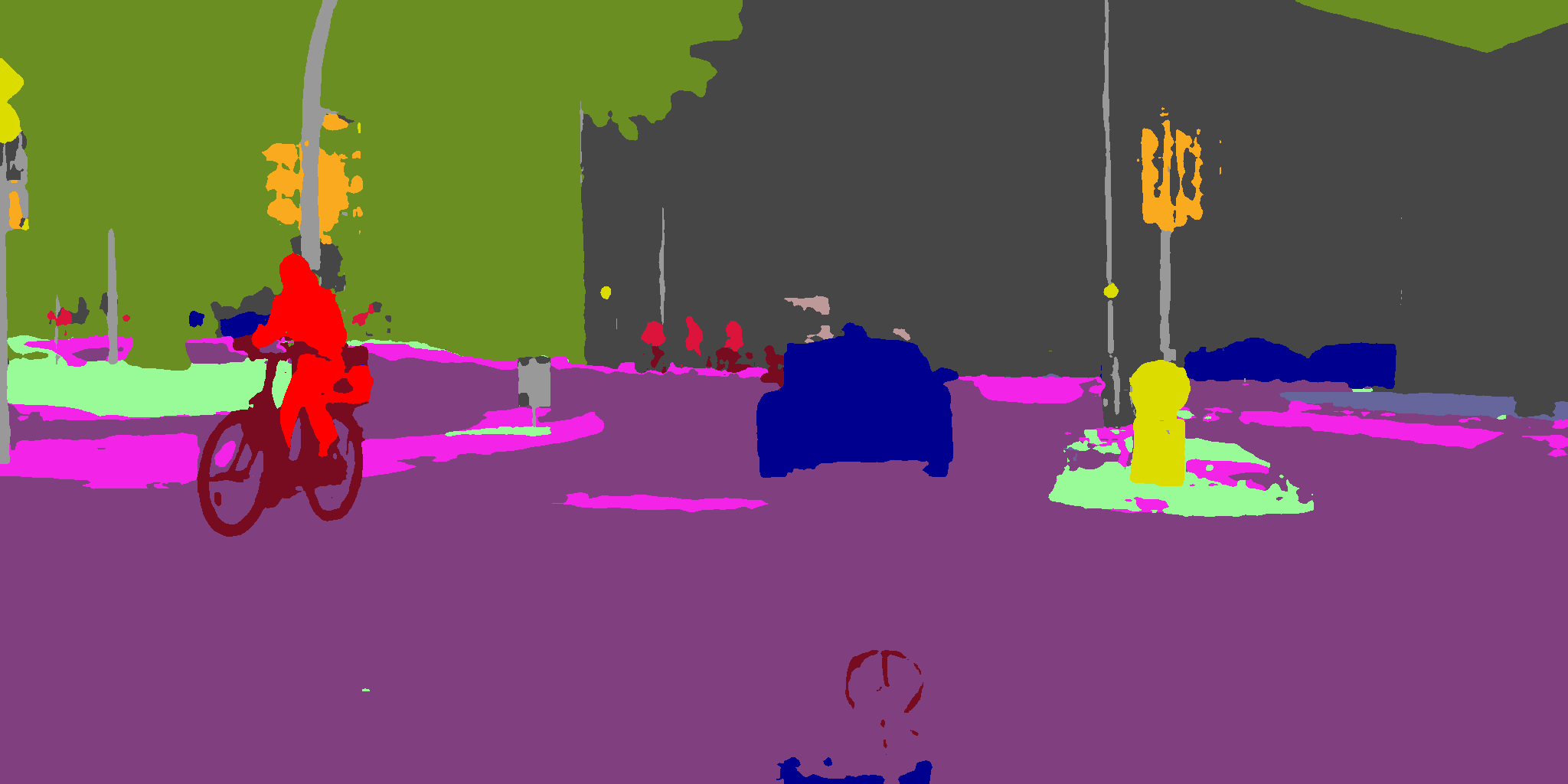} & \includegraphics[width=0.15\textwidth]{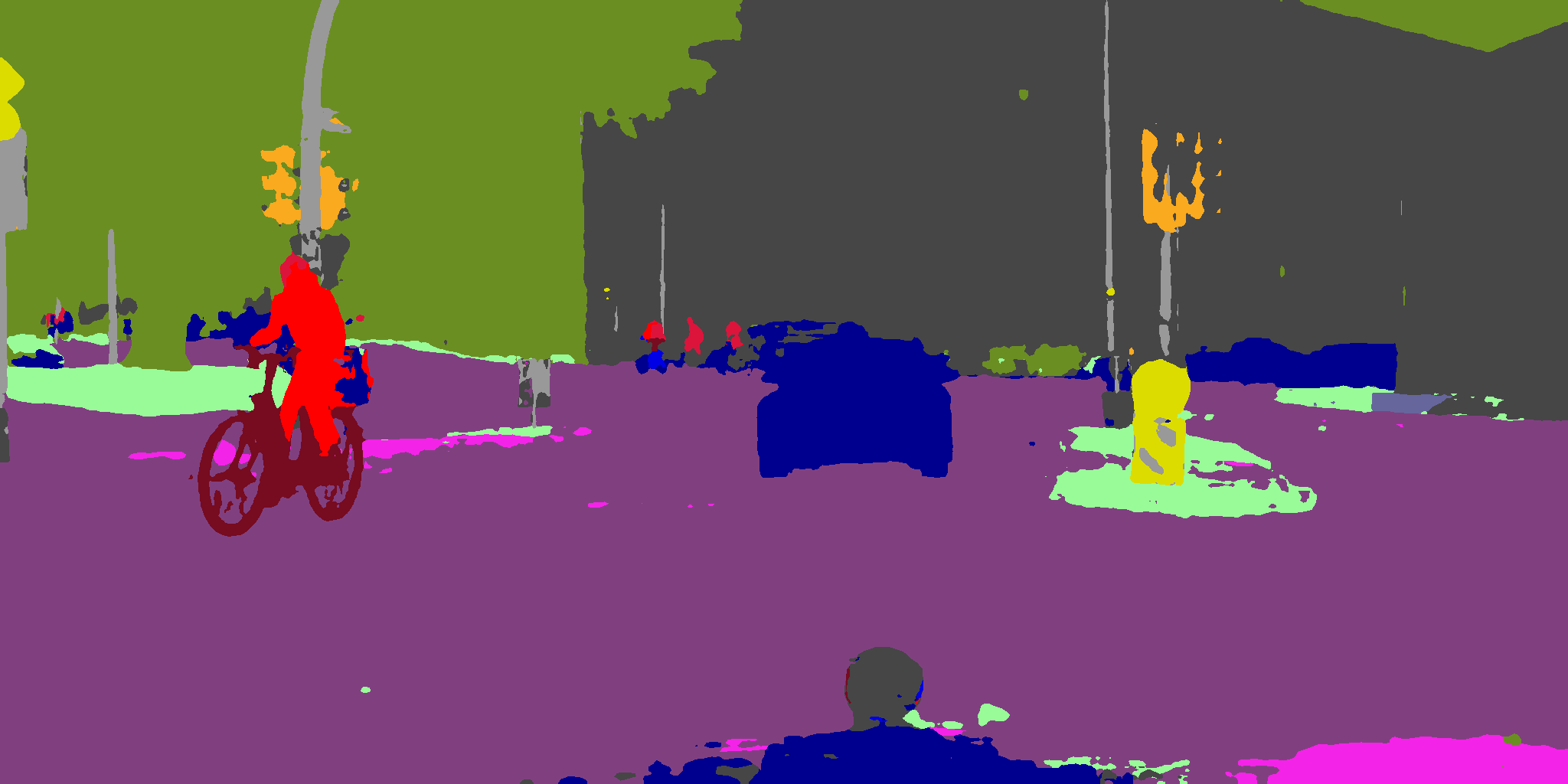} & \includegraphics[width=0.15\textwidth]{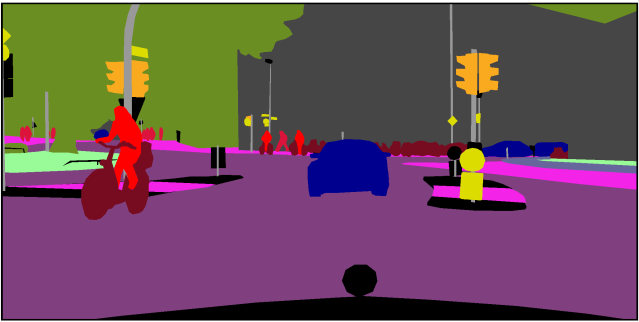}\\
        
        \includegraphics[width=0.15\textwidth]{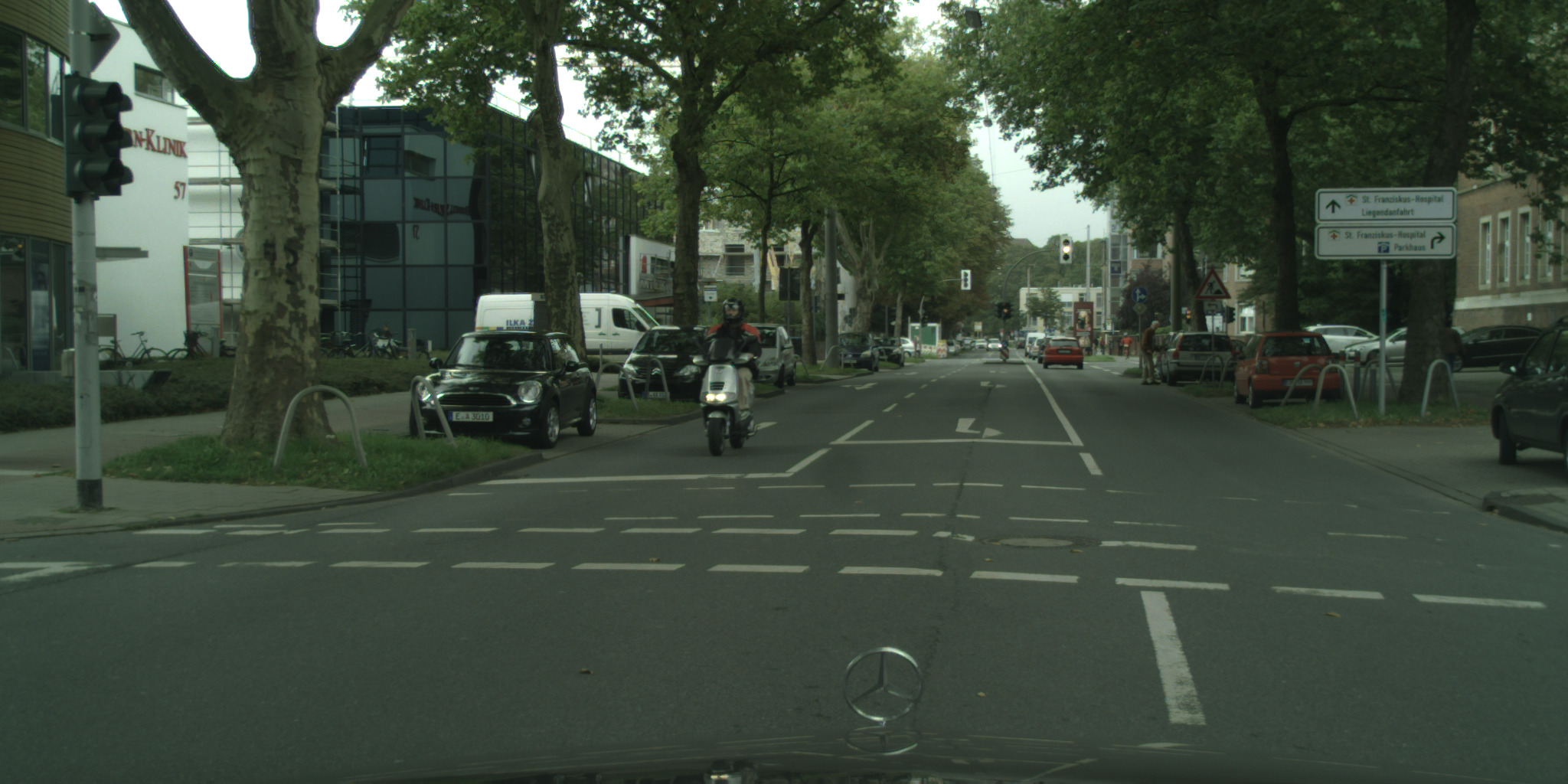} & \includegraphics[width=0.15\textwidth]{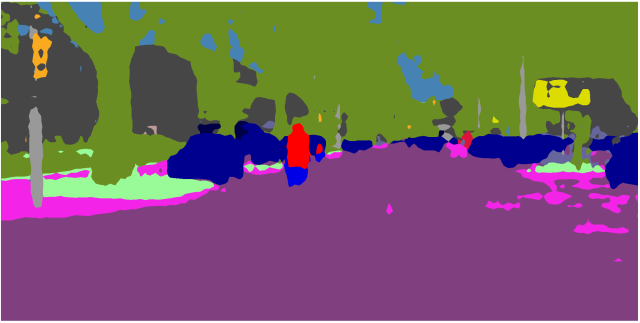} & \includegraphics[width=0.15\textwidth]{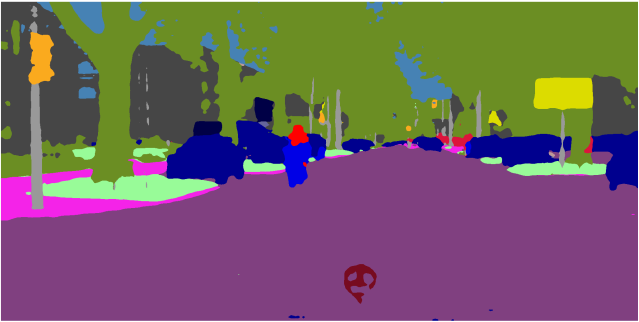} & \includegraphics[width=0.15\textwidth]{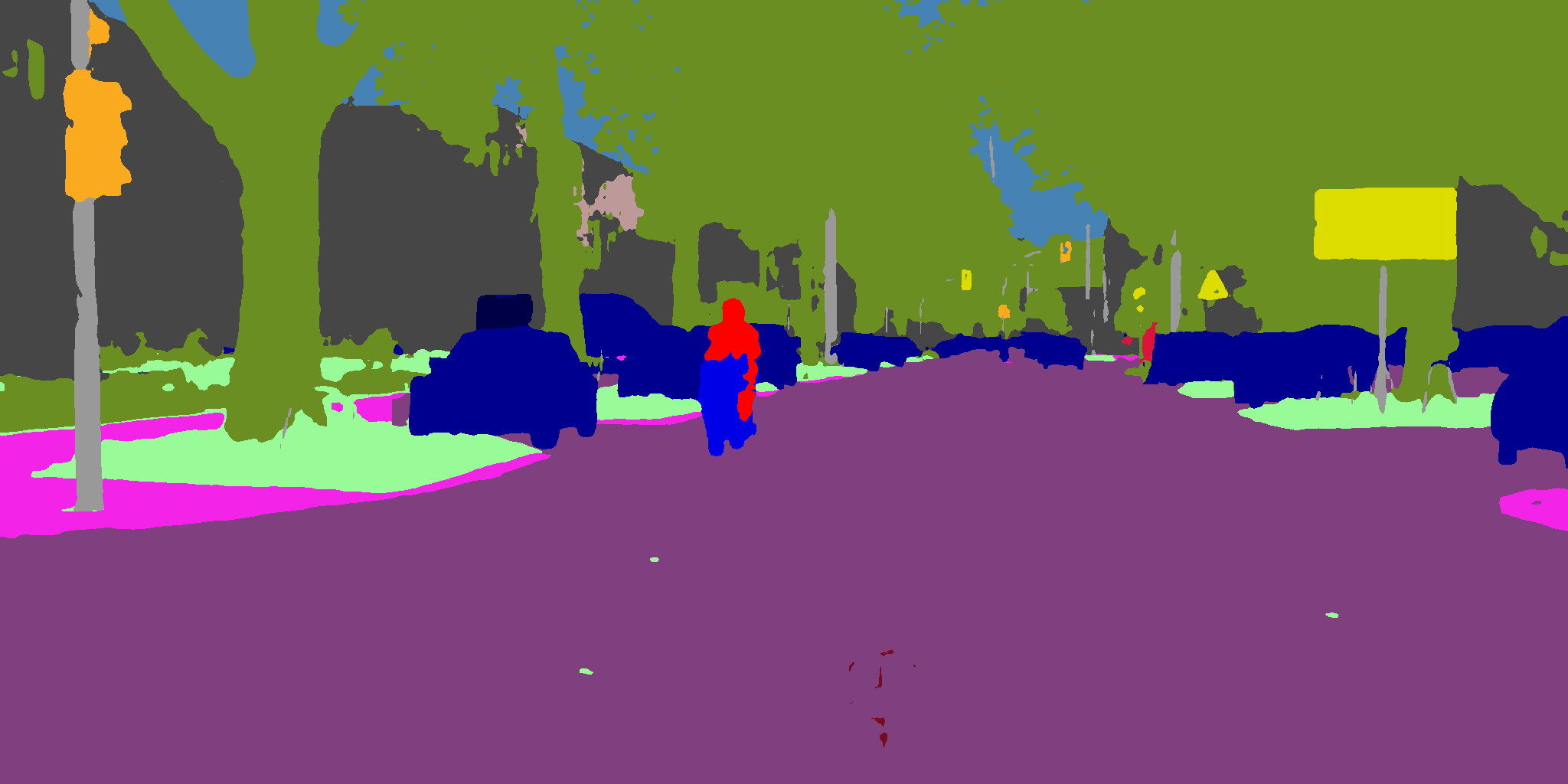} & \includegraphics[width=0.15\textwidth]{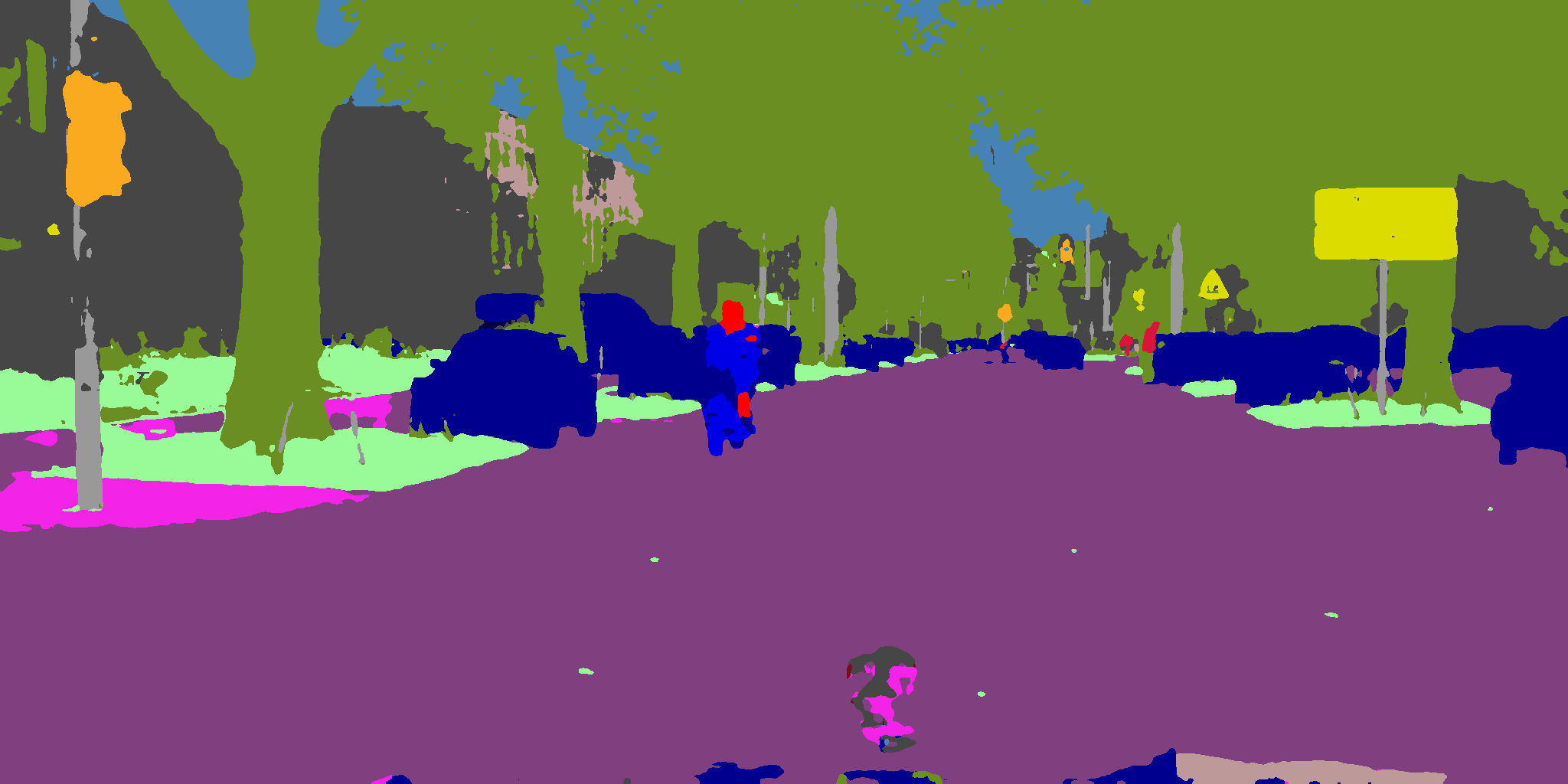} & \includegraphics[width=0.15\textwidth]{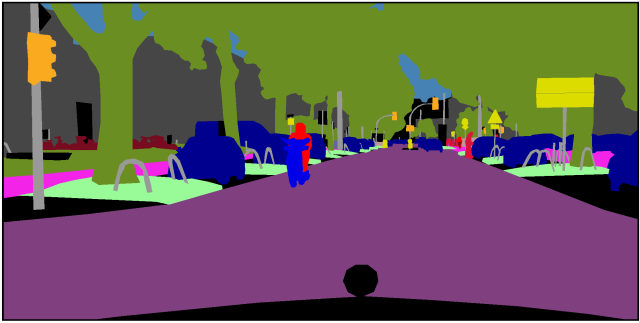}\\
        
        \includegraphics[width=0.15\textwidth]{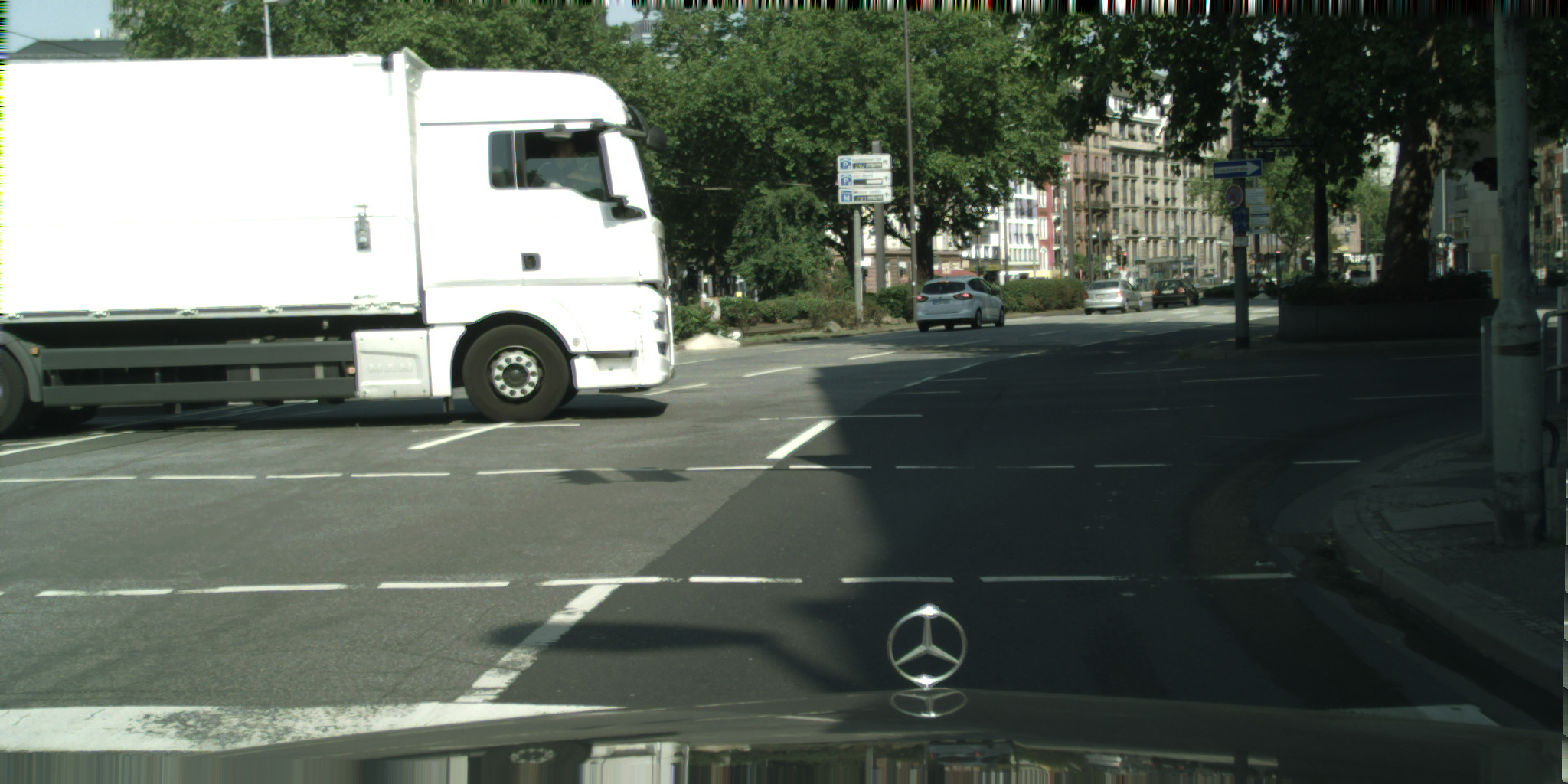} & \includegraphics[width=0.15\textwidth]{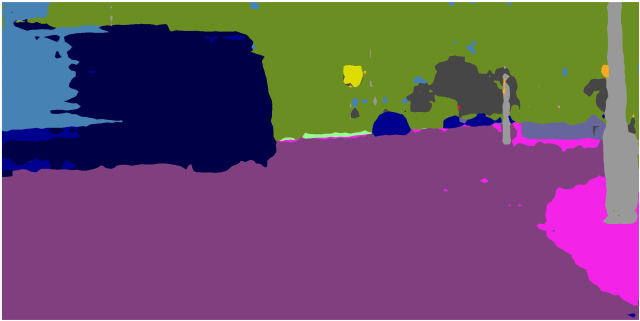} & \includegraphics[width=0.15\textwidth]{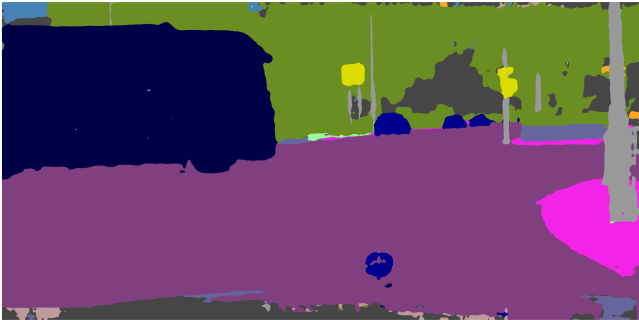} & \includegraphics[width=0.15\textwidth]{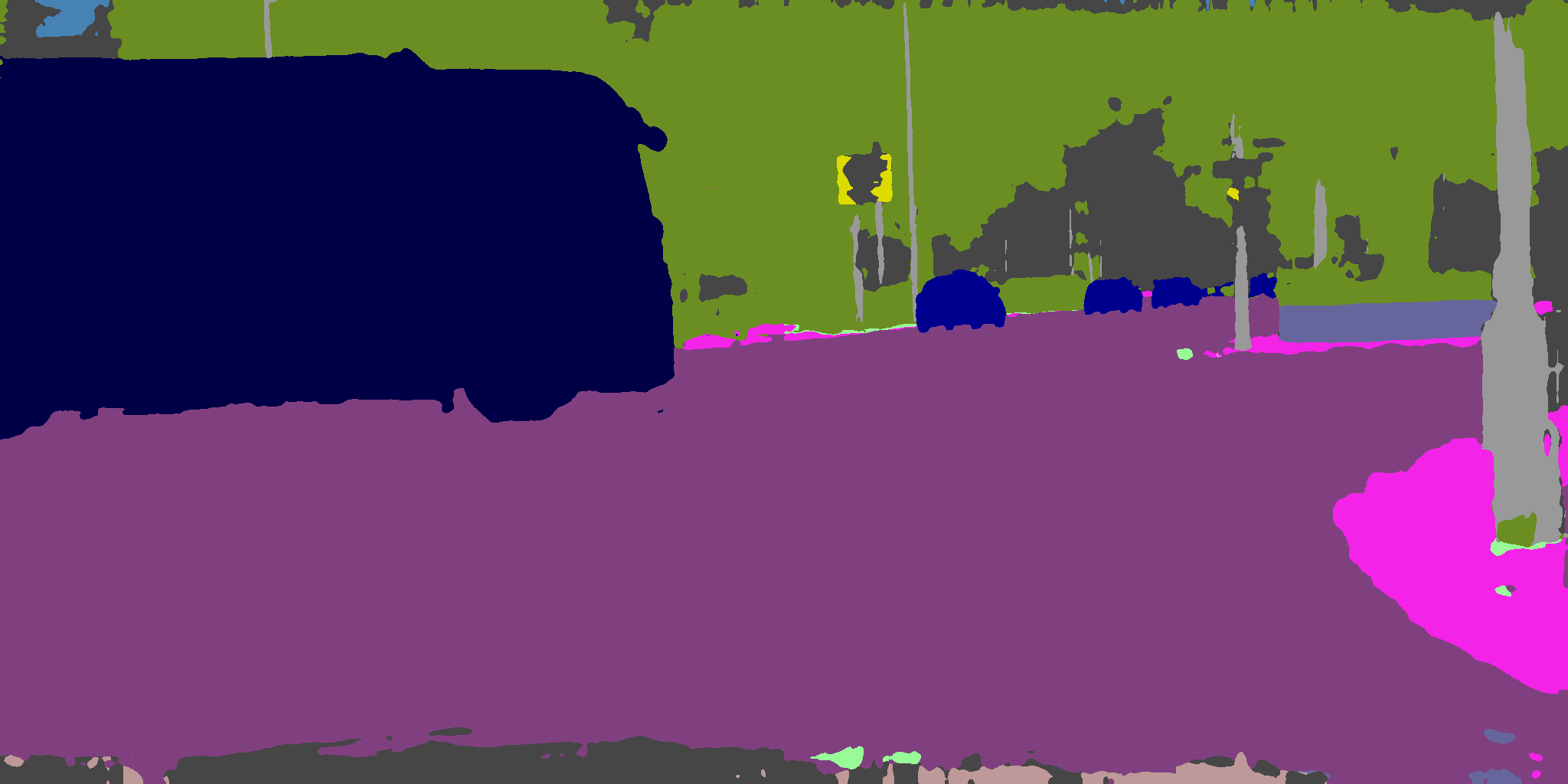} & \includegraphics[width=0.15\textwidth]{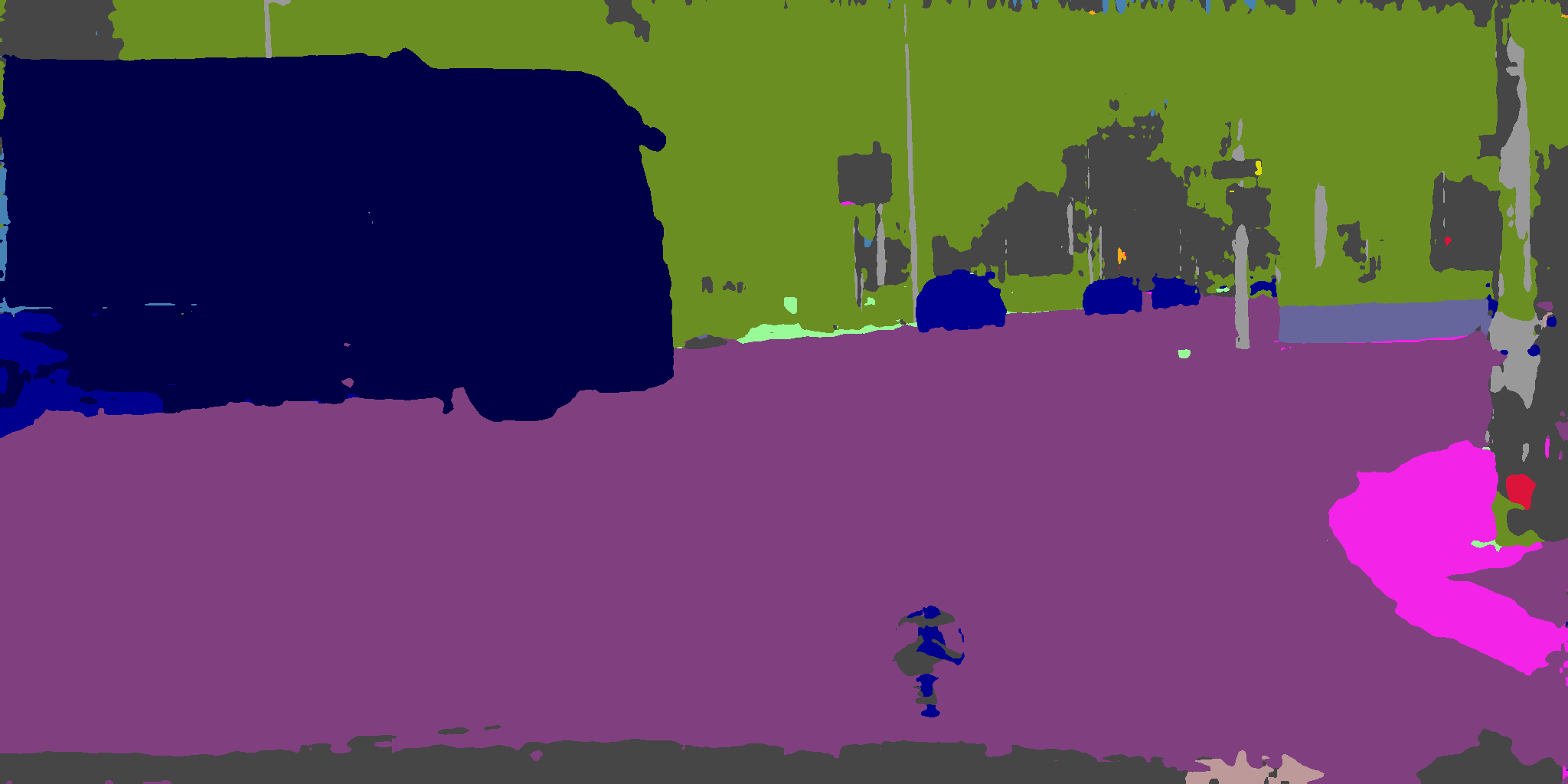} & \includegraphics[width=0.15\textwidth]{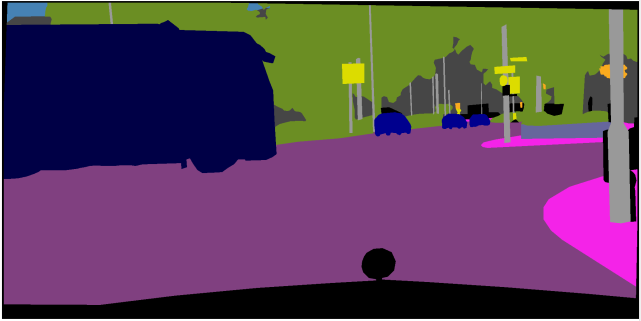} \\

        {\small Target Image} & {\small ASM} & {\small CACDA} & {\small HRDA} & {\small \makecell{HRDA+\\\method (Ours)}} & {\small GT}\\
    \end{tabular}%
    }
    \caption{\textbf{Qualitative results of segmentation maps.} We compare the segmentation maps from different UDA methods on the \gtoc benchmark.}
    \label{tab:segmaps}
\end{figure*}